\documentclass[10pt,twocolumn,letterpaper]{article}

\usepackage{adjustbox}

\usepackage[pagenumbers]{cvpr} 
\usepackage{graphicx}
\usepackage{float}
\usepackage{subcaption}
\usepackage{adjustbox} 

\usepackage{multirow}

\usepackage{listings}
\lstdefinelanguage{json}{
    morestring=[b]",%
    morestring=[s]{'}{'},%
    morecomment=[l]{//},%
    morecomment=[s]{/*}{*/},%
    morekeywords={true,false,null},%
    sensitive=false,%
    alsoletter={:},%
    moredelim=[l][\color{black}\ttfamily]{"},
    moredelim=[s][\color{black}\ttfamily]{:}{,},
    moredelim=[s][\color{black}\ttfamily]{\{}\},
}

\usepackage{xcolor}

\definecolor{cvprblue}{rgb}{0.21,0.49,0.74}
\usepackage[pagebackref,breaklinks,colorlinks,allcolors=cvprblue]{hyperref}

\def\paperID{20291} 
\def\confName{CVPR}
\def\confYear{2026}

\title{\textsc{RecruitView}: A Multimodal Dataset for Predicting Personality and Interview Performance for Human Resources Applications}

\author{
Amit~Kumar~Gupta\textsuperscript{1}\thanks{Equal contribution as first authors.}\, \thanks{Corresponding author: \texttt{amit.gupta@jaipur.manipal.edu}}
\quad
Farhan~Sheth\textsuperscript{1}\footnotemark[1]
\quad
Hammad~Shaikh\textsuperscript{1}
\quad
Dheeraj~Kumar\textsuperscript{1}
\\
Angkul~Puniya\textsuperscript{1}
\quad
Deepak~Panwar\textsuperscript{1}
\quad
Sandeep~Chaurasia\textsuperscript{1}
\quad
Priya~Mathur\textsuperscript{2}
\\[0.3em]
\textsuperscript{1}Manipal University Jaipur, India
\qquad
\textsuperscript{2}Poornima Institute of Engineering \& Technology, India
\\[0.3em]
{\tt\small \{amit.gupta, deepak.panwar, sandeep.chaurasia\}@jaipur.manipal.edu}
\\[-0.2em]
{\tt\small \{farhan.219310185, hammad.229301534\}@muj.manipal.edu}
\\[-0.2em]
{\tt\small \{dheeraj.229301593, angkul.23FE10CAI00309\}@muj.manipal.edu}
\\[-0.2em]
{\tt\small priya.mathur@poornima.org}
}

\begin{document}
\maketitle





\begin{abstract}
Automated personality and soft skill assessment from multimodal behavioral data remains challenging due to limited datasets and methods that fail to capture geometric structure inherent in human traits. We introduce \textit{\textsc{RecruitView}}, a dataset of 2,011 naturalistic video interview clips from 300+ participants with 27,000 pairwise comparative judgments across 12 dimensions: Big Five personality traits, overall personality score, and six interview performance metrics. To leverage this data, we propose Cross-Modal Regression with Manifold Fusion (\texttt{CRMF}), a geometric deep learning framework that explicitly models behavioral representations across hyperbolic, spherical, and Euclidean manifolds. \texttt{CRMF} employs geometry-specific expert networks to capture hierarchical trait structures, directional behavioral patterns, and continuous performance variations simultaneously. An adaptive routing mechanism dynamically weights expert contributions based on input characteristics. Through principled tangent space fusion, \texttt{CRMF} achieves superior performance while training 40--50\% fewer trainable parameters than large multimodal models. Extensive experiments demonstrate that \texttt{CRMF} substantially outperforms the selected baselines, achieving up to 11.4\% improvement in Spearman correlation and 6.0\% in concordance index. Our \textit{\textsc{RecruitView}} dataset is publicly available at \texttt{\small\url{https://huggingface.co/datasets/AI4A-lab/RecruitView}}.
\end{abstract}

\section{Introduction}
\label{sec:intro}


Interviews are integral to hiring, coaching, and clinical evaluation. Judgments hinge on subtle behaviors distributed across what candidates say, how they speak, and how they present visually. As video interviewing scales, computational assessment must read these signals coherently rather than in isolation. Estimating personality traits and interview performance from short video responses is a multimodal problem spanning vision, speech, and language. Interviews rely on complementary lexical, prosodic, and visual cues, therefore computational models must capture these complementary signals without discarding their structure.

General-purpose LMMs (MiniCPM-o~\cite{hu2024minicpm}, VideoLLaMA2~\cite{cheng2024videollama2}, Qwen2.5-Omni~\cite{chu2024qwen2}) offer breadth but are not tuned for fine-grained social inference. Conventional fusion maps all modalities to a single Euclidean latent via concatenation or vanilla attention, ignoring modality-specific geometry. A single latent geometry limits representational adequacy.

Progress is also constrained by supervision: existing datasets are noisy, weakly controlled, and often not domain-specific; they typically lack multi-trait personality and interview-related metrics, instead relying on direct scalar ratings that are sensitive to scale use and inter-rater variability. To address this, we introduce \textsc{RecruitView}—\textbf{R}ecorded \textbf{E}valuations of \textbf{C}andidate \textbf{R}esponses for \textbf{U}nderstanding \textbf{I}ndividual \textbf{T}raits—a multimodal interview corpus of 2{,}011 clips from more than 300 sessions, each aligned to one of 76 questions. Clinical psychologists provided about 27,000 pairwise comparisons between answers to the same prompt, which we convert into continuous scores for 12 targets, namely the Big Five traits~\cite{mccrae1992five}, an overall personality score, and six interview performance metrics, using a nuclear-norm-regularized multinomial logit model. This protocol reduces rater calibration biases and yields reliable regression labels. 

We propose \textbf{C}ross-Modal \textbf{R}egression with \textbf{M}anifold \textbf{F}usion (\texttt{CRMF}), a geometry-aware framework that projects fused multimodal features to hyperbolic, spherical, and Euclidean spaces, processes each with a geometry-specific expert, and aggregates them through input-adaptive routing with geometry-aware attention and tangent-space fusion. This design preserves manifold consistency while enabling input-conditioned combination for multi-target regression. 


\noindent\textbf{The contribution of this work} is fourfold: (i) \textsc{RecruitView}, a multimodal interview dataset with psychometrically grounded labels derived from pairwise judgments mapped to continuous scores, covering 12 targets across personality and performance; (ii) \texttt{CRMF}, a principled geometry-aware fusion framework that learns in hyperbolic, spherical, and Euclidean spaces; (iii) an adaptive routing and geometry-aware attention mechanism with tangent-space fusion for input-conditioned combination of geometric experts; and (iv) a comprehensive evaluation demonstrating consistent gains over recent LMM baselines on all metrics.

\section{Related Works}
\label{sec:LR}

\subsection{Personality and Behavioral Assessment}

Automated personality and performance assessment has relied on datasets such as ChaLearn~\cite{ponce2016chalearn} and POM~\cite{park2014pom}, which advanced the field but remain limited by controlled settings and narrow labeling scopes. Later efforts like YouTube Personality~\cite{biel2011facetube} and Interview2Personality~\cite{song2020interview2personality} moved toward more naturalistic or interview-style data yet still suffer from smaller scale, scripted responses, and subjective absolute ratings. However, \textsc{RecruitView} is an in-the-wild interview dataset, where labels are obtained via pairwise comparisons, yielding consistent continuous scores. Beyond the Big Five and an overall personality index, \textsc{RecruitView} annotates performance dimensions (e.g., confidence, communication), enabling joint modeling of personality and interview behavior.


\subsection{Multimodal Fusion for Behavioral Analysis}
Multimodal fusion has been extensively studied for affective computing and personality recognition. Early work focused on feature-level concatenation~\cite{baltrusaitis2018multimodal} or attention-based aggregation~\cite{tsai2019multimodal,zadeh2017tensor}. Transformer-based architectures have recently dominated this space, with methods like MULT~\cite{tsai2019multimodal} employing cross-modal attention for temporal alignment. However, these approaches operate entirely in Euclidean space, potentially missing important geometric structure in behavioral data.

Recent large multimodal models have shown remarkable zero-shot and few-shot capabilities. MiniCPM-o~\cite{hu2024minicpm} employs an end-to-end training paradigm with modality-adaptive modules, while VideoLLaMA2~\cite{cheng2024videollama2} introduces spatial-temporal visual token compression for efficient video understanding. Qwen2.5-Omni~\cite{chu2024qwen2} extends text-centric LLMs with native audio-visual understanding through cross-attention fusion. Despite their general-purpose success, these models lack task-specific inductive biases for personality assessment and are not optimized for capturing the geometric properties of behavioral traits.

\subsection{Geometric Deep Learning}
Geometric deep learning extends neural networks to non-Euclidean domains. Hyperbolic neural networks~\cite{ganea2018hyperbolic,chami2019hyperbolic} leverage the exponentially growing capacity of hyperbolic space to model hierarchical data, showing benefits for tree-structured tasks and knowledge graph reasoning. Spherical networks~\cite{cohen2018spherical,esteves2018learning} operate on the unit sphere, naturally suited for directional data and rotational equivariance. Recent work has explored mixed-curvature spaces~\cite{gu2019learning,skopek2020mixed} that combine multiple geometries, though primarily for representation learning rather than multimodal fusion.

Manifold-valued neural networks~\cite{brooks2019riemannian,lou2020neural} perform operations directly on Riemannian manifolds, ensuring geometric consistency. However, these methods have seen limited application in behavioral analysis. Our work is the first to systematically leverage multiple geometric manifolds for multimodal behavioral assessment with learned adaptive fusion.

\subsection{Mixture-of-Experts Architectures}
Mixture-of-experts (MoE) models~\cite{shazeer2017outrageously,fedus2022switch} decompose complex tasks into specialized sub-networks selected by a gating function. Traditional MoE aims for sparse activation to increase model capacity efficiently. Recent work has extended MoE to multimodal settings~\cite{mustafa2022multimodal,xin2025i2moe} and to geometric spaces~\cite{lou2020neural}. However, existing geometric MoE methods typically focus on sparsity for computational efficiency rather than complementary geometric reasoning. Our routing mechanism differs fundamentally: rather than encouraging specialization, we promote diversity to leverage complementary geometric views of behavioral data, with all experts contributing to the final prediction through learned weighting.

\section{\textsc{RecruitView}}
\label{sec:dataset}

To satisfy the critical necessity of a psychometrically robust dataset to analyze multimodal interview performance and personality, we introduce \textsc{RecruitView}. This novel dataset comprises 2,011 video segments sampled from 331 distinct interview sessions. Specifically developed to facilitate training and testing on demanding, human-centered traits, it provides robust, continuous labels on 12 distinct targets. The dataset's key contribution is in the form of an annotation method through pairwise ratings by clinical psychologists to mitigate rater bias and provide robust, continuous scores. The following sections describe the deliberate process of developing it in stages from stimulus creation to final form of data.

\subsection{Data Collection}
The creation of \textsc{RecruitView} followed a two-phase approach: creation of a broad-based question repository to be used as a prompter, and the procurement of video replies by a diverse group of respondents.

Our dataset has as its base a specially selected pool of queries crafted to elicit responses suitable to human resources evaluation and personality rating. For this purpose, we carried out an exhaustive compilation exercise from a variety of sources. We went through public domain material on interview preparation by market leaders, analyzed frequently posed queries on professional networking platforms, and closely interacted with clinical psychologists. This made the queries relevant not only in the context of professional hiring but also well suited to exploring the underlying Big Five personality dimensions.

This procedure gave rise to 76 standard interview questions (the full list of questions is available in Appendix~\ref{app:questions}). The questions were then systematically and in a balanced manner sorted into 15 individual sets for convenience in the collection of data. There were five individual questions in each set and a typical opening question (``Introduce yourself" or ``Tell me about yourself"), thereby establishing a comparable baseline in the majority of interviews but facilitating greater query variety within the dataset.

Participants were students from various Manipal universities, who responded via a custom web platform and were randomly assigned to one of 15 question sets. Interviews were recorded in diverse in-the-wild settings (e.g., classrooms, private residences). Implementation details of the platform are provided in Appendix~\ref{app:websites}.

Data collection and use followed institutional ethical approval processes; detailed ethics, consent, and risk-mitigation discussion is provided in Appendix~\ref{app:ethics}.


\subsection{Dataset Annotation}
\label{sec:annotation}

\subsubsection{Annotation Protocol}
To ensure consistency and psychometric reliability in subjective evaluations, we employed a pairwise comparison protocol inspired by prior multimodal labeling frameworks such as the ChaLearn dataset \cite{ponce2016chalearn, chen2016overcoming}. Instead of assigning absolute scores, clinical psychologists were presented with two clips responding to the same interview question and asked to identify which participant better demonstrated a target attribute, for example, “Who appears more confident?” Annotators could also indicate a tie when both clips were judged equivalent. This comparative design minimizes calibration bias, reduces inter-rater variability, and enhances reliability in perceptual assessments \cite{thurstone1927law}. The protocol was applied across twelve target dimensions covering the Big Five personality traits (Openness, Conscientiousness, Extraversion, Agreeableness, and Neuroticism), Overall Personality, and six interview or performance-related metrics: Interview Score, Answer Score, Speaking Skills, Confidence Score, Facial Expression, and Overall Performance. In total, approximately 27,310 pairwise judgments were collected, forming the basis for deriving continuous and psychometrically grounded labels.

\subsubsection{Model Selection and Multinomial Logit (MNL)}
We evaluated several frameworks for converting pairwise judgments into continuous labels, including Elo rating \cite{glickman1999rating}, Bradley-Terry-Luce (BTL) \cite{bradley1952rank, luce1959individual}, TrueSkill \cite{herbrich2007trueskill}, and Glicko-2 \cite{glickman1999parameter}. While these models are widely used in ranking applications, they either assume strong independence across traits or lack convex formulations with clear identifiability guarantees. After empirical comparison (results in Appendix~\ref{app:labeling_comparison}), we selected the Multinomial Logit (MNL) \cite{negahban2018learning} model with nuclear norm regularization, which offered both strong theoretical grounding and robust empirical performance on our dataset.

\noindent Each video $j$ is associated with a latent utility $\theta_j \in \mathbb{R}$. For a comparison between clips $j_1$ and $j_2$, the MNL model defines the probability that $j_1$ is preferred as
\begin{equation}
    \Pr\{ j_1 \succ j_2 \} = 
    \frac{\exp(\theta_{j_1})}{\exp(\theta_{j_1}) + \exp(\theta_{j_2})}
\end{equation}

\noindent Letting $X^{(i)}$ denote the design matrix for comparison $i$ and $y_i \in \{0,1\}$ its observed outcome, the normalized log-likelihood across $n$ comparisons is
\begin{equation}
    \mathcal{L}(\Theta) = \frac{1}{n} \sum_{i=1}^{n} 
    \Big( y_i \langle \Theta, X^{(i)} \rangle - 
    \log\big(1 + \exp(\langle \Theta, X^{(i)} \rangle)\big) \Big)
\end{equation}
where $\Theta \in \mathbb{R}^{N \times T}$ is the matrix of utilities across $N$ videos and $T=12$ targets.

\subsubsection{Nuclear Norm Regularization and Optimization}
Recovering utilities requires regularization to address limited sampling and correlations across traits. We therefore estimate $\Theta$ by solving the convex program
\begin{equation}
    \hat{\Theta} = \arg\min_{\Theta \in \Omega} 
    \Big[ -\alpha \, \mathcal{L}(\Theta) + 
    \lambda \| L^{1/2}\Theta \|_{*} \Big]
\end{equation}
where $\|\cdot\|_{*}$ denotes the nuclear norm \cite{fazel2002matrix}, $L$ is the Laplacian of the comparison graph, and $\Omega$ constrains identifiability (e.g., centering utilities). The Laplacian-induced nuclear norm encourages low-rank structure while respecting the blockwise nature of the pairwise comparisons (same-question groups).

We solve this convex program using first-order proximal methods with singular value shrinkage \cite{cai2010singular}, implemented in \texttt{cvxpy}\footnote{\url{https://www.cvxpy.org/}} with an \texttt{SCS}\footnote{\url{https://www.cvxgrp.org/scs/}} solver. Step sizes are adapted with the Barzilai--Borwein \cite{barzilai1988two} rule to accelerate convergence. The resulting $\hat{\Theta}$ provides continuous, psychometrically grounded labels for all 12 target dimensions.

\subsection{Data Format and Structure}
\label{subsec:data_format}

The \textsc{RecruitView} dataset comprises 2,011 multimodal samples, each representing a candidate's response to one of 76 interview questions. Each sample is structured to facilitate comprehensive multimodal analysis through three primary components:

\begin{itemize}[leftmargin=*, noitemsep, topsep=2pt]
    \item \textbf{Video}: High-resolution recordings stored in compressed MP4 format at 30 FPS. The dataset’s average video duration is approximately 30 seconds.
    
    \item \textbf{Audio}: High-fidelity audio tracks extracted from videos (mono channel).
    
    \item \textbf{Transcript}: Verbatim speech-to-text transcriptions automatically generated using Whisper-large-v3\footnote{\url{https://huggingface.co/openai/whisper-large-v3}} \cite{radford2023robust}.
\end{itemize}

\noindent\textbf{Metadata and Annotations:} All annotations and metadata are organized in a structured JSON format. Each entry contains a unique identifier, video filename, interview question, quality indicators (video quality, duration category), user number and the 12 continuous target scores derived from the pairwise comparison protocol (see Appendix~\ref{app:json_sample} for a complete sample entry). This unified structure ensures seamless integration across modalities while maintaining data privacy and facilitating reproducible research workflows.

\subsection{Task and Metrics}
The primary task enabled by the \textsc{RecruitView} dataset is multimodal regression. Given a video clip of a candidate's response, the goal is to predict the 12 continuous scores corresponding to their personality traits and interview performance. Models leverage the modalities available from the data: visual (video frames), auditory (speech acoustics), and linguistic (transcribed text).

\noindent The 12 target variables for prediction are divided into the following two categories:

\noindent\textbf{\textit{Personality Traits Metrics:}} These are based on the widely accepted Five-Factor Model of personality, with an additional overall score.

\noindent\textbf{1. Openness (O):} Measures imagination, creativity, and intellectual curiosity. Individuals high in openness are often inventive and enjoy new experiences.

\noindent\textbf{2. Conscientiousness (C):} Assesses self-discipline, organization, and goal-directed behavior. High conscientiousness is associated with being hardworking and reliable. 

\noindent\textbf{3. Extraversion (E):} Reflects sociability, assertiveness, and emotional expressiveness. Extroverts tend to be outgoing and energized by social interaction.

\noindent\textbf{4. Agreeableness (A):} Indicates compassion, cooperativeness, and trustworthiness. Agreeable individuals are often helpful and empathetic.

\noindent\textbf{5. Neuroticism (N):} Pertains to emotional stability. Individuals high in neuroticism tend to experience negative emotions like anxiety and stress more frequently.

\noindent\textbf{6. Overall Personality:} A holistic assessment of the participant's perceived personality, derived from the combination of the Big Five traits.

\noindent\textbf{\textit{Performance Metrics:}} These six metrics evaluate key competencies and behaviors exhibited during an interview response.

\noindent\textbf{7. Interview Score:} An overall score assessing the holistic quality of the participant's interview segment.

\noindent\textbf{8. Answer Score:} Evaluates the content of the response, including its relevance to the question, coherence, and structured thinking.

\noindent\textbf{9. Speaking Skills:} Assesses vocal characteristics such as clarity, pace, tone, and the avoidance of filler words.

\noindent\textbf{10. Confidence Score:} Measures the degree of self-assurance projected by the participant through both verbal and non-verbal cues (e.g., posture, eye contact, vocal tone).

\noindent\textbf{11. Facial Expression:} Quantifies the extent to which the participant uses facial expressions to convey emotion and engagement.

\noindent\textbf{12. Overall Performance:} A comprehensive evaluation of the candidate's performance in the clip, integrating all other performance factors.



\subsection{Data Statistics}
\begin{figure}[!t]
    \centering
    \includegraphics[width=\linewidth]{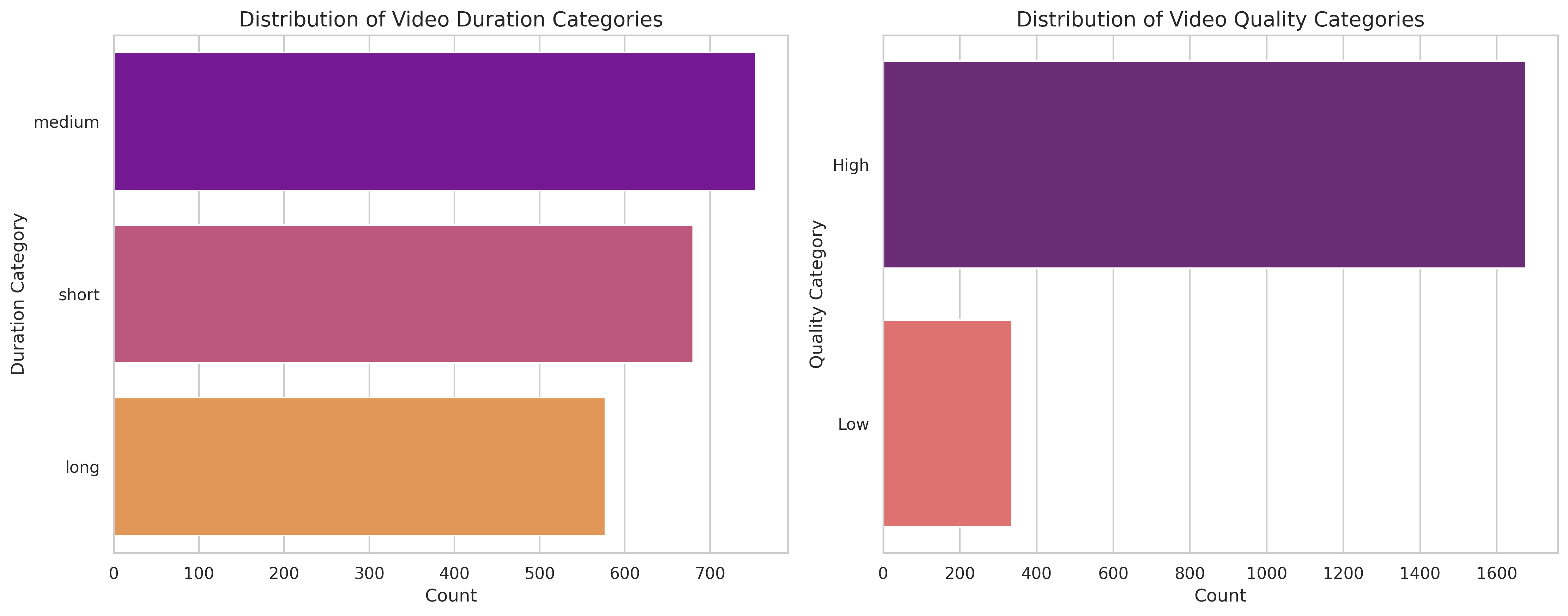}
    \caption{Data distribution by duration and video quality.}
    \label{fig:video_dur_qua}
\end{figure}

The \textsc{RecruitView} corpus comprises 2,011 video segments, sourced from over 300 unique participants responding to a bank of 76 curated interview questions. The dataset's foundation is a set of approximately 27,000 pairwise judgments provided by clinical psychologists. A key design characteristic is its ``in-the-wild" data collection via a custom-built web-based platform. This methodology encouraged participation in naturalistic settings, resulting in significant variability in lighting conditions, background environments, and audio quality. This inherent diversity ensures high ecological validity, a critical feature for developing robust models that can generalize beyond controlled laboratory conditions. Figure~\ref{fig:video_dur_qua} provides a summary of the video duration and quality categories, illustrating the distribution of these factors across the corpus. 

\begin{figure}[!t]
    \centering
    \includegraphics[width=\linewidth]{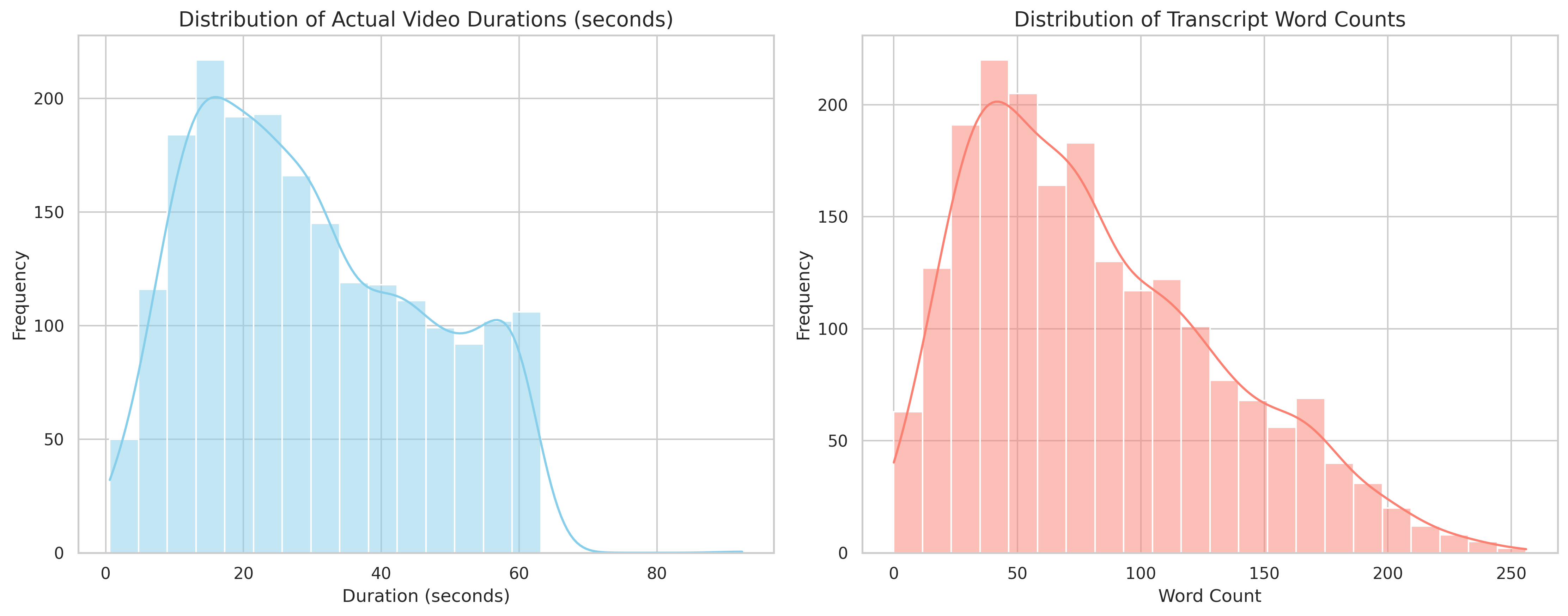}
    \caption{Data distribution by durations and transcript word count.}
    \label{fig:video_dur_text_count}
\end{figure}

To provide a more granular view of the dataset's temporal and linguistic composition, Figure~\ref{fig:video_dur_text_count} illustrates the distributions for video segment duration and the corresponding transcript word counts. The video durations (Figure~\ref{fig:video_dur_text_count}, left) follow a right-skewed distribution, with a primary mode around 20-30 seconds and a secondary mode near 60 seconds, reflecting the natural variance in response length. The transcript word counts (Figure~\ref{fig:video_dur_text_count}, right) are similarly distributed, with most responses falling between 40 and 115 words. This confirms that the dataset captures a wide range of response styles, from brief, concise answers to more detailed, elaborate ones.

Detailed summary statistics for video durations and transcript word counts are provided in Appendix~\ref{app:data_stats}, confirming the temporal and linguistic diversity of the dataset.

\subsubsection{Correlation Analysis}
\label{subsec:correlation}

To understand the relationships among the 12 target dimensions in \textsc{RecruitView}, we computed Spearman rank correlations across all 2,011 video clips. 

\begin{figure}[!t]
\centering
    \begin{subfigure}[t]{0.48\linewidth}
        \centering
        \includegraphics[width=\linewidth]{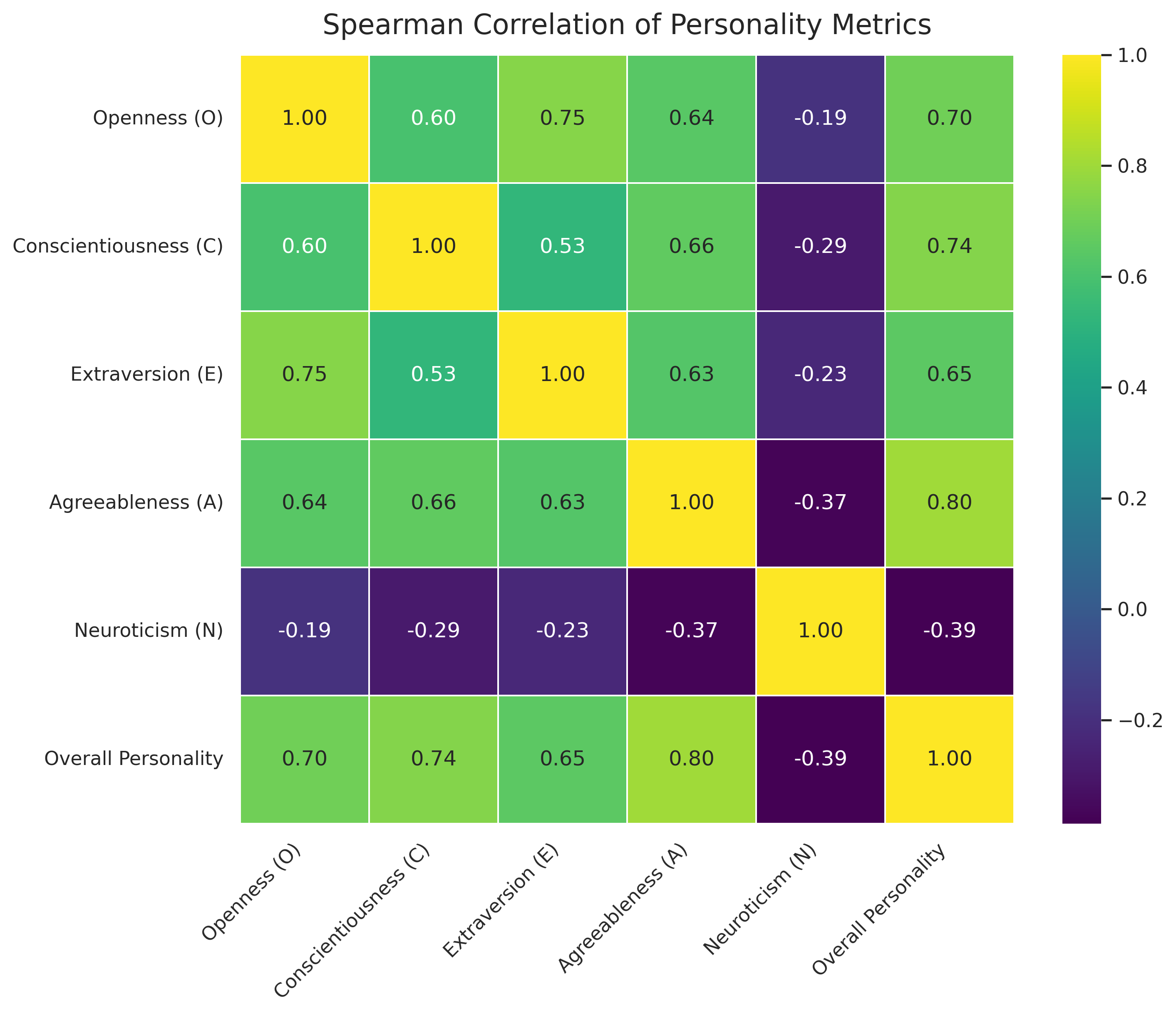}
        \caption{Spearman’s $\rho$ correlation matrix for personality trait metrics.}
        \label{fig:corr_traits_personality}
    \end{subfigure}
    \hfill
    \begin{subfigure}[t]{0.48\linewidth}
        \centering
        \includegraphics[width=\linewidth]{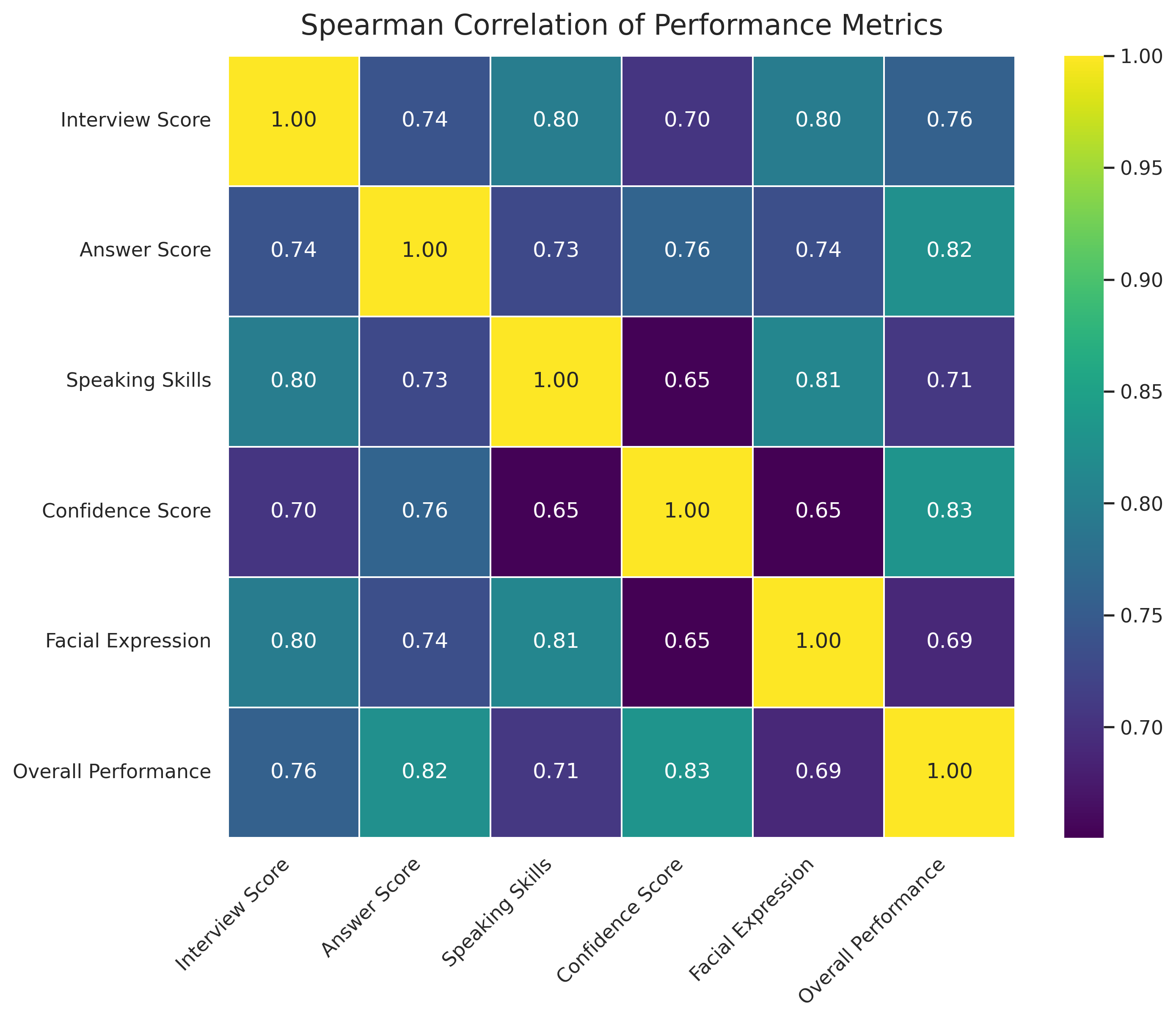}
        \caption{Spearman’s $\rho$ correlation matrix for performance metrics.}
        \label{fig:corr_perf_performance}
    \end{subfigure}
    \caption{Spearman correlation between various metrics.}
    \label{fig:spearman_correlation}
\end{figure}

\paragraph{Personality Trait Metrics.}
Figure~\ref{fig:corr_traits_personality} shows positive correlations between Overall Performance and Openness ($\rho = 0.70$), Conscientiousness ($\rho = 0.74$), Extraversion ($\rho = 0.65$), with Agreeableness strongest ($\rho = 0.80$); Neuroticism is negative ($\rho = -0.39$). This pattern aligns with theory: interpersonal warmth (Agreeableness) is most salient, while organization and intellectual engagement (Conscientiousness, Openness) also contribute.

\paragraph{Performance Metrics.}
Figure~\ref{fig:corr_perf_performance} shows moderate–strong positive correlations across all metrics. Confidence has the strongest association ($\rho = 0.83$), followed by Answer Score ($\rho = 0.82$) and Interview Score ($\rho = 0.76$). Facial Expressions and Speaking Skills are also substantial ($\rho = 0.69$, $\rho = 0.71$). Overall, stronger perceived personality aligns with higher confidence, more expressive nonverbal behavior, and better-structured, well-delivered responses.


The complete $12 \times 12$ Spearman correlation matrix with all pairwise relationships is provided in Appendix~\ref{subsec:full_correlation} for comprehensive reference.

\subsubsection{Metrics Statistics}
\label{subsec:metrics_stats}
We analyzed the statistical properties of the 12 continuous target labels derived from the MNL model. The distributions are all normalized with near-zero means. A complete, detailed statistical analysis of all 12 metrics, including their implications, is provided in Appendix~\ref{app:metrics_stats_analysis}.
Most metrics exhibit significant leptokurtosis (heavy tails) and asymmetric skew. For instance, Speaking Skills ($\rho \approx -0.86$) and Overall Performance ($\rho \approx -0.75$) are negatively skewed, while Answer Score ($\rho \approx 0.35$) is positively skewed. This prevalence of outliers and non-normality strongly motivates our methodological choices: (1) the use of a robust \textbf{Huber loss} for training to mitigate the influence of extreme outliers, and (2) the prioritization of \textbf{rank-based correlation metrics} (e.g., Spearman's $\rho$) for evaluation, which are insensitive to this skew. 

\noindent\textbf{Outlier Treatment via Soft Winsorization.} To address extreme outliers while preserving data structure, we apply mild soft winsorization. Values within $\pm1.5\sigma$ pass through unchanged, while values beyond this threshold are smoothly compressed toward $\pm3\sigma$ using tanh-based soft clipping: $\text{clip}(x) = \text{sign}(x) \cdot (\theta + s \cdot \tanh((|x| - \theta) / s))$ for $|x| > \theta$, where $\theta = 1.5$ and $s = 1.5$. This smooth transition prevents extreme values from dominating gradient updates while maintaining differentiability and rank ordering, improving convergence without discarding informative variance.




\section{Methodology}
\label{sec:methodology}

\subsection{Problem Formulation}

We address the problem of predicting multiple continuous attributes from multimodal behavioral data. Given a video clip containing visual frames $\mathbf{V} \in \mathbb{R}^{T \times H \times W \times 3}$, audio waveform $\mathbf{A} \in \mathbb{R}^{L}$, and transcript text $\mathbf{T}$, our goal is to predict a vector of target attributes $\mathbf{y} \in \mathbb{R}^{K}$ representing personality traits and performance scores. Formally, we learn a function $f: (\mathbf{V}, \mathbf{A}, \mathbf{T}) \rightarrow \mathbf{y}$ that captures the complex relationships between multimodal behavioral cues and target attributes.

Traditional approaches assume all representations reside in Euclidean space, employing linear transformations and standard neural operations. However, behavioral data exhibits diverse geometric properties: personality traits form hierarchical taxonomies (Big Five domains comprising specific facets), behavioral cues show directional relationships (facial expressions oriented in specific directions), and performance metrics often vary continuously. To capture this rich structure, we propose explicitly modeling representations across multiple geometric manifolds, each encoding different relational properties of the data.

\subsection{CRMF Architecture Overview}

\begin{figure*}[!t]
    \centering
    \includegraphics[width=\linewidth]{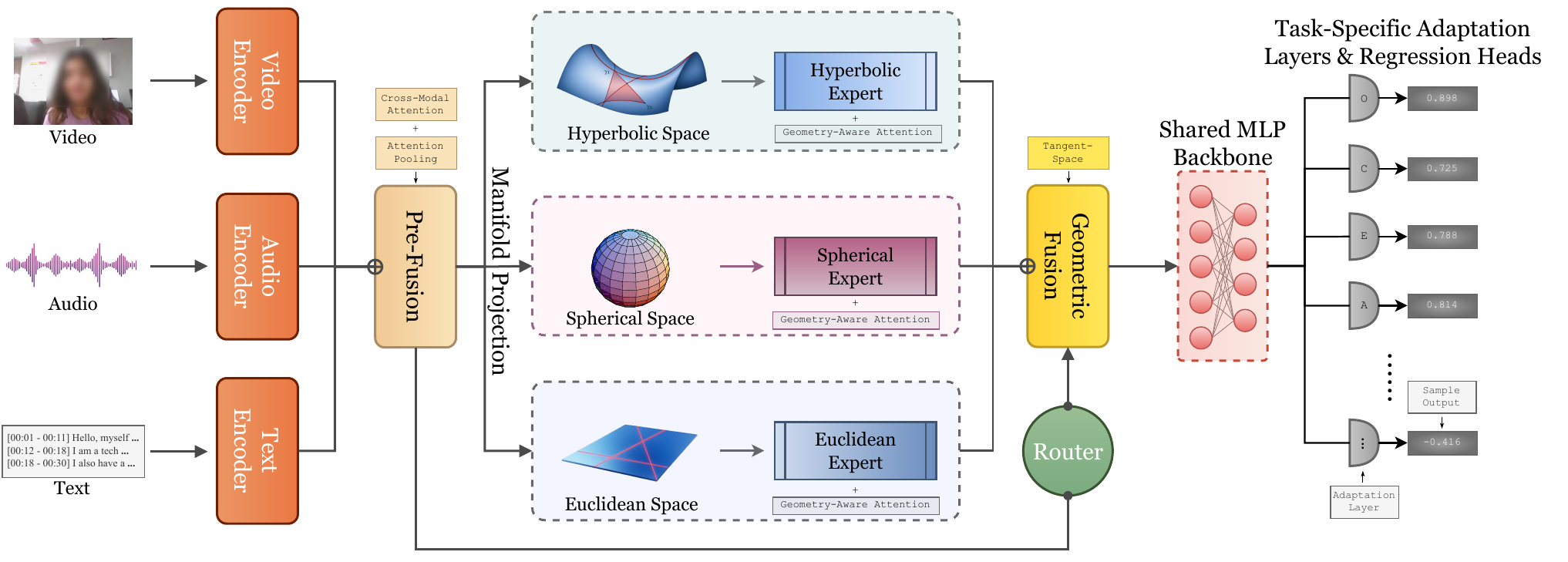}
    \caption{Overview of the \texttt{CRMF} architecture. Multimodal encoders extract features from video, audio, and text. Pre-fusion integrates modalities through cross-modal attention. The manifold projection layer maps features to hyperbolic, spherical, and Euclidean spaces. Geometry-specific experts process each manifold representation with intra-manifold attention. A learned router dynamically weights expert outputs. Finally, geometric fusion combines representations in a shared tangent space for multi-target prediction.}
    \label{fig:architecture}
\end{figure*}

The \texttt{CRMF} framework consists of six core components: (1) modality-specific encoders that extract representations from each input channel, (2) a pre-fusion module that performs early cross-modal integration, (3) manifold projection layers that map features to three geometric spaces, (4) geometry-specific expert networks that process each manifold representation, (5) an adaptive routing mechanism that learns optimal geometric combination strategies, and (6) a geometric fusion module that integrates multi-geometry representations for final prediction. Figure~\ref{fig:architecture} illustrates the complete architecture.

The key insight behind \texttt{CRMF} is that different aspects of behavioral assessment benefit from different geometric inductive biases. Hyperbolic geometry naturally encodes hierarchical trait structures, spherical geometry captures directional behavioral patterns, and Euclidean geometry models continuous performance variations. By processing features through all three geometries and learning to combine them adaptively, \texttt{CRMF} can capture the full complexity of behavioral data. A detailed description of the \texttt{CRMF} framework’s formulation and architecture is provided in Appendix~\ref{app:crmf-detailed}.

\subsection{Multimodal Encoding}

We employ pretrained encoders for each modality: DeBERTa-v3-base~\cite{he2021deberta} for text, Wav2Vec2~\cite{baevski2020wav2vec}/HuBERT~\cite{hsu2021hubert} for audio, and VideoMAE~\cite{tong2022videomae}/TimeSformer~\cite{bertasius2021spacetime} for video. We fine-tune the last few layers of each encoder while keeping earlier layers frozen for parameter efficiency. For video, we apply a sophisticated temporal modeling pipeline consisting of BiLSTM, multi-head attention, and 1D convolution to capture rich temporal dynamics before pre-fusion. All modality encoders output representations with unified dimension $d_{model} = 768$. Full details are provided in the Appendix~\ref{app:multimodal-encoding}.

\subsection{Pre-Fusion Module}

The pre-fusion module performs early integration of multimodal features through cross-modal attention. We concatenate encoded features from all modalities and add learnable modality embeddings:
\begin{equation}
    \mathbf{H}_{cat} = [\mathbf{H}_t; \mathbf{H}_a; \mathbf{H}_v] + \mathbf{E}_{mod}
\end{equation}
where $\mathbf{E}_{mod} \in \mathbb{R}^{3 \times d}$ contains unique embeddings for text, audio, and video. A multi-layer transformer encoder processes the concatenated sequence, enabling rich cross-modal interactions. We employ learned attention pooling to obtain a fixed-dimensional clip-level representation $\mathbf{z}_{pre} \in \mathbb{R}^{d}$.

\subsection{Manifold Projection and Geometric Experts}

We project the fused representation $\mathbf{z}_{pre}$ onto three geometric manifolds using learned linear projections followed by geometry-specific mappings:

\noindent\textbf{Hyperbolic Space:} We use the Poincaré ball model $\mathbb{B}^d_c$ with curvature $c = 1.0$. Points are mapped via exponential map: $\exp_{\mathbf{0}}^c(\mathbf{v}) = \tanh(\sqrt{c}\|\mathbf{v}\|) \frac{\mathbf{v}}{\sqrt{c}\|\mathbf{v}\|}$, where $\mathbf{v} = \mathbf{W}_h \mathbf{z}_{pre}$.

\noindent\textbf{Spherical Space:} The unit sphere $\mathbb{S}^{d-1}$ is parameterized through $L_2$ normalization: $\mathbf{x}_s = \frac{\mathbf{W}_s \mathbf{z}_{pre}}{\|\mathbf{W}_s \mathbf{z}_{pre}\| + \epsilon}$.

\noindent\textbf{Euclidean Space:} Standard linear projection: $\mathbf{x}_e = \mathbf{W}_e \mathbf{z}_{pre}$.

Each manifold representation is processed by a specialized expert network designed to respect the underlying geometry. The hyperbolic expert uses Möbius transformations in gyrovector space, the spherical expert operates via tangent space mappings with exponential/logarithmic maps, and the Euclidean expert uses standard feed-forward layers. All experts have multiple layers and residual connections. Detailed formulations are available in the Appendix~\ref{app:geometric-experts}.

\subsection{Geometry-Aware Attention}

To further refine expert outputs, we apply intra-manifold attention that respects geometric structure. For each geometry, we compute attention in its respective tangent space, which is Euclidean and enables standard multi-head attention operations. The attended representations are then mapped back to their respective manifolds. 

\subsection{Adaptive Token Routing}

The router learns to weight expert outputs based on input characteristics. Given $\mathbf{z}_{pre}$, a lightweight MLP predicts routing weights:
\begin{equation}
    \mathbf{r} = \text{softmax}(\mathbf{W}_r^{(2)} \sigma(\mathbf{W}_r^{(1)} \mathbf{z}_{pre} + \mathbf{b}_r^{(1)}) + \mathbf{b}_r^{(2)})
\end{equation}
where $\mathbf{r} \in \Delta^{K-1}$ contains weights for the $K=3$ experts. To encourage diverse geometry utilization, we apply entropy regularization: $\mathcal{L}_{entropy} = -\lambda_{ent} \sum_{i=1}^{K} r_i \log r_i$. A negative value encourages high entropy, promoting complementary geometric views.

\subsection{Geometric Fusion}

The fusion module combines expert outputs from different manifolds by first mapping all to a shared tangent space. For hyperbolic and spherical outputs, we use logarithmic maps; Euclidean output requires no conversion. The fusion operates via routing-weighted average:
\begin{equation}
    \mathbf{z}_{fusion} = r_h \mathbf{v}_h + r_s \mathbf{v}_s + r_e \mathbf{v}_e
\end{equation}
followed by a refinement network producing $\mathbf{z}_{refined}$. This strategy is equivalent to first-order Fréchet mean approximation on the product manifold.

\subsection{Multi-Task Prediction Head}

The prediction head maps the fused representation to target attributes using a shared MLP backbone followed by lightweight task-specific adaptation layers for each of the $K=12$ targets. This parameter-efficient design balances expressiveness and efficiency.

\subsection{Training Objective}

Our training objective combines multiple loss components through adaptive balancing:
\begin{equation}
    \mathcal{L}_{total} = \sum_{i=1}^{N} \beta_i \mathcal{L}_i
\end{equation}
where components include Huber regression loss ($\delta = 1.0$), correlation boosting loss, covariance alignment loss, and auxiliary routing regularization losses. The weights $\beta_i$ are learned adaptively using inverse variance weighting combined with learned parameters. Full details are in the Appendix~\ref{app:training-objective}.






\section{Experimental Setup}


\subsection{Implementation Details}
We train \texttt{CRMF} using AdamW~\cite{loshchilov2018adamw} with component-specific learning rates. We use batch size 4 with gradient accumulation over 8 steps (effective batch size 32) and OneCycleLR scheduling with 15\% warmup. Training runs for 30 epochs with early stopping on validation Spearman correlation (patience 5). Text is tokenized with max length 512, audio resampled to 16kHz, and video sampled at 16 FPS with 16 frames per clip resized to $224 \times 224$.

\subsection{Baselines and Evaluation}

We compare against three recent large multimodal models: MiniCPM-o 2.6 (8B)~\cite{hu2024minicpm}, VideoLLaMA2.1-AV (7B)~\cite{cheng2024videollama2}, and Qwen2.5-Omni (7B)~\cite{chu2024qwen2}, all fine-tuned on our task. We evaluate using Spearman's $\rho$, Kendall's $\tau$-b, Concordance Index (C-Index), Pearson's $r$, and MSE. Metrics are computed per-target and macro-averaged for overall performance.

\section{Results}

\subsection{Overall Performance}
Table~\ref{tab:macro-results} presents aggregate results averaged across all 12 target attributes. \texttt{CRMF} variants substantially outperform all baseline models across correlation and ranking metrics. The best \texttt{CRMF} configuration (VideoMAE + Wav2Vec2) achieves Spearman $\rho = 0.5682$, representing an 11.4\% relative improvement over the strongest baseline (MiniCPM-o's 0.5102). Similar gains are observed for Kendall's $\tau$-b (14.9\% improvement) and concordance index (6.0\% improvement).

\noindent\textbf{Parameter Efficiency:} The performance gains are particularly remarkable given \texttt{CRMF}'s parameter efficiency. Our framework (VideoMAE + Wav2Vec2 configuration) contains 408M parameters total, with 172M trainable during fine-tuning. In contrast, baseline LMMs fine-tune substantially more parameters: MiniCPM-o ($\sim$340M), VideoLLaMA2 ($\sim$300M), Qwen2.5-Omni ($\sim$320M). Despite training 40-50\% fewer \emph{trainable} parameters, \texttt{CRMF} achieves superior performance, demonstrating that task-specific geometric inductive biases provide more effective learning signals than simply leveraging larger pretrained models.

Comparing encoder choices, VideoMAE generally outperforms TimeSformer, suggesting that masked autoencoding provides better video representations for this task. For audio, Wav2Vec2 and HuBERT show comparable performance, with Wav2Vec2 having a slight edge on correlation metrics.

\begin{table}[!t]
\scriptsize
\centering
\setlength{\tabcolsep}{5pt}
\renewcommand{\arraystretch}{1.15}
\begin{adjustbox}{width=\linewidth}
\begin{tabular}{l | cccc}
\toprule
    \textbf{Model} &
    \textbf{Spearman $\rho$} &
    \textbf{Kendall $\tau$-b} &
    \textbf{C-index} & 
    \textbf{Pearson $r$} \\
\midrule
    MiniCPM-o 2.6 (8B) & 0.5102 & 0.3541 & 0.6779 & 0.4935 \\
    VideoLLaMA2.1-AV (7B) & 0.5002 & 0.3498 & 0.6778 & 0.4802 \\
    Qwen2.5-Omni (7B) & 0.4882 & 0.3378 & 0.6658 & 0.4682 \\
\midrule
    \texttt{CRMF} (VMAE + w2v2) & \textbf{0.5682} & \textbf{0.4069} & \textbf{0.7183} & 0.5475 \\
    \texttt{CRMF} (VMAE + HuB) & 0.5645 & 0.4020 & 0.7158 & \textbf{0.5481} \\
    \texttt{CRMF} (TimeS + w2v2) & 0.5581 & 0.3980 & 0.7103 & 0.5394 \\
    \texttt{CRMF} (TimeS + HuB) & 0.5664 & 0.4020 & 0.7148 & 0.5547 \\
\bottomrule
\end{tabular}
\end{adjustbox}
\caption{Macro-averaged performance across all targets. \texttt{CRMF} variants consistently outperform baseline LMMs. Best results are bolded.}
\label{tab:macro-results}
\end{table}

\subsection{Per-Dimension Analysis}

Per-trait personality assessment results (Table~\ref{tab:per-personality-results} in Appendix) show \texttt{CRMF} demonstrates strong performance across all Big Five dimensions. Openness shows the strongest \texttt{CRMF} performance ($\rho = 0.6384$), representing a 13.4\% improvement over the best baseline. Conscientiousness, Extraversion, and Agreeableness exhibit moderate but consistent improvements (8-13\% gains). Neuroticism presents the most challenging prediction task, though \texttt{CRMF} still improves upon baselines by 24.5\%.

Per-dimension performance assessment results (Table~\ref{tab:per-performance-results} in Appendix) detail results for six performance-related attributes. Interview Score and Answer Score show the strongest absolute performance ($\rho > 0.62$ and $\rho > 0.59$), with 9-12\% improvements over baselines. Speaking Skills and Confidence Score achieve moderate but consistent improvements (10-16\% gains). Overall Performance benefits most from \texttt{CRMF} ($\rho = 0.6521$), with 9.1\% improvement.

\subsection{Ablation Studies}

Table~\ref{tab:ablation-main} presents key ablation results. Using only fusion (no \texttt{CRMF} framework), such as simple concatenation ($\rho = 0.4441$) or weighted averaging ($\rho = 0.4664$), causes severe degradation (21.8\% and 17.9\% drops), demonstrating that naive fusion strategies fail to capture complex multimodal relationships. 

\begin{table}[!t]
\scriptsize
\centering
\setlength{\tabcolsep}{4pt}
\renewcommand{\arraystretch}{1.15}
\begin{adjustbox}{width=\linewidth}
\begin{tabular}{ll | ccc}
\toprule
    \textbf{Component} & \textbf{Variant} & \textbf{$\rho$} & \textbf{$\tau$-b} & \textbf{C-idx} \\
\midrule
    \multicolumn{2}{l|}{\textit{Full \texttt{CRMF} Model}} & \textit{0.5682} & \textit{0.4069} & \textit{0.7183} \\
\midrule
    \multirow{2}{*}{\textbf{Fusion Only}} & Simple Concatenation & 0.4441 & 0.2903 & 0.6151 \\
    & Weighted Average & 0.4664 & 0.3078 & 0.6239 \\
\midrule
    \multirow{3}{*}{\textbf{Geometry}} & Hyperbolic Only & 0.5080 & 0.3611 & 0.6980 \\
    & Spherical Only  & 0.5338 & 0.3808 & 0.7054 \\
    & Euclidean Only & 0.5284 & 0.3753 & 0.7001 \\
\midrule
    \textbf{Routing} & No Router (Uniform) & 0.5209 & 0.3688 & 0.6994 \\
\midrule
    \multirow{3}{*}{\textbf{Modality}} & Video Only (VMAE) & 0.4516 & 0.2974 & 0.6521 \\
    & Audio Only (Wav2Vec2) & 0.3792 & 0.2461 & 0.6245 \\
    & Text Only (DeBERTa) & 0.4247 & 0.2806 & 0.6429 \\
\bottomrule
\end{tabular}
\end{adjustbox}
\caption{Key ablation study results demonstrating the contribution of \texttt{CRMF} components. All experiments use VideoMAE+Wav2Vec2.}
\label{tab:ablation-main}
\end{table}

Using only a single geometric space consistently underperforms: Hyperbolic-only achieves $\rho = 0.5080$ (10.6\% drop), Spherical-only $\rho = 0.5338$ (6.1\% drop), and Euclidean-only $\rho = 0.5284$ (7.0\% drop). No single geometry matches the full model, confirming that different geometric spaces capture complementary information.

Removing the router and using uniform weights ($\rho = 0.5209$) causes substantial degradation (8.3\% drop), confirming that adaptive weighting based on input characteristics is crucial.

Single modality analysis reveals video provides the strongest unimodal signal ($\rho = 0.4516$), followed by text ($\rho = 0.4247$) and audio ($\rho = 0.3792$). Crucially, even the best unimodal result is substantially lower than any multimodal configuration. The leap from video-only to full \texttt{CRMF} represents a 25.8\% improvement, underscoring that behavioral traits are expressed through complex interplay of cues across modalities.

Additional ablation results for two-geometry combinations, pre-fusion variants, prediction head architectures, and loss functions are provided in the Appendix (Table~\ref{tab:crmf-ablation-full}).
                                               
\section{Conclusion}

We introduced \textsc{RecruitView}, a multimodal corpus for personality and interview analysis with continuous, psychometrically grounded labels derived from pairwise judgments. Building on this resource, we proposed \texttt{CRMF}, a geometry-aware regression framework that fuses audio, video, and text using manifold-specific attention and adaptive routing. On \textsc{RecruitView}, \texttt{CRMF} surpasses strong multimodal baselines, raising macro Spearman correlation to 0.568 and concordance index to 0.718, while using fewer trainable parameters. Ablations validate the benefits of multi-geometry fusion and routing, and show clear gaps between multimodal and unimodal variants. Limitations include moderate dataset scale, short clips, potential residual annotator bias and label noise despite calibration, and limited demographic diversity, which may constrain external validity. Future work will broaden populations and conditions, extend to longer multi-turn interactions, and integrate stronger self-supervised priors, target-wise manifold selection, and causal analyses, alongside real-time inference and human-in-the-loop calibration.


\section*{Funding}
This work was funded by the Manipal Research Board (MRB) Research Grant, Letter No. \textbf{DoR/MRB/2023/SG-08}.

\section*{Acknowledgment}
We thank Manipal University Jaipur (MUJ) for providing research infrastructure, computing resources, and institutional support that made this work possible. We are grateful to the Office of Research and the Manipal Research Board (MRB) for guidance and administrative assistance. We also thank our colleagues for constructive discussions and feedback.

\section*{Data and Code Availability}
The \textsc{RecruitView} dataset and the \texttt{CRMF} implementation are publicly available. The dataset is hosted on Hugging Face at \url{https://huggingface.co/datasets/AI4A-lab/RecruitView} and mirrored on GitHub at \url{https://github.com/AI4A-lab/RecruitView}. The \texttt{CRMF} framework code is available at \url{https://github.com/AI4A-lab/CRMF}.


{
    \small
    \bibliographystyle{ieeetr}
    \bibliography{main}

@String(ECCV= {Eur. Conf. Comput. Vis.})

@String(NIPS= {Adv. Neural Inform. Process. Syst.})

@String(ECCV  = {ECCV})

@String(NIPS  = {NeurIPS})

@article{hu2024minicpm,
  title={Minicpm: Unveiling the potential of small language models with scalable training strategies},
  author={Hu, Shengding and Tu, Yuge and Han, Xu and He, Chaoqun and Cui, Ganqu and Long, Xiang and Zheng, Zhi and Fang, Yewei and Huang, Yuxiang and Zhao, Weilin and others},
  journal={arXiv preprint arXiv:2404.06395},
  year={2024}
}

@article{cheng2024videollama2,
  title={Videollama 2: Advancing spatial-temporal modeling and audio understanding in video-llms},
  author={Cheng, Zesen and Leng, Sicong and Zhang, Hang and Xin, Yifei and Li, Xin and Chen, Guanzheng and Zhu, Yongxin and Zhang, Wenqi and Luo, Ziyang and Zhao, Deli and others},
  journal={arXiv preprint arXiv:2406.07476},
  year={2024}
}

@article{chu2024qwen2,
  title={Qwen2.5-omni technical report},
  author={Xu, Jin and Guo, Zhifang and He, Jinzheng and Hu, Hangrui and He, Ting and Bai, Shuai and Chen, Keqin and Wang, Jialin and Fan, Yang and Dang, Kai and others},
  journal={arXiv preprint arXiv:2503.20215},
  year={2025}
}

@article{baltrusaitis2018multimodal,
  title={Multimodal machine learning: A survey and taxonomy},
  author={Baltru{\v{s}}aitis, Tadas and Ahuja, Chaitanya and Morency, Louis-Philippe},
  journal={IEEE transactions on pattern analysis and machine intelligence},
  volume={41},
  number={2},
  pages={423--443},
  year={2018},
  publisher={IEEE}
}

@inproceedings{tsai2019multimodal,
  title={Multimodal transformer for unaligned multimodal language sequences},
  author={Tsai, Yao-Hung Hubert and Bai, Shaojie and Liang, Paul Pu and Kolter, J Zico and Morency, Louis-Philippe and Salakhutdinov, Ruslan},
  booktitle={Proceedings of the conference. Association for computational linguistics. Meeting},
  volume={2019},
  pages={6558},
  year={2019}
}

@article{zadeh2017tensor,
  title={Tensor fusion network for multimodal sentiment analysis},
  author={Zadeh, Amir and Chen, Minghai and Poria, Soujanya and Cambria, Erik and Morency, Louis-Philippe},
  journal={arXiv preprint arXiv:1707.07250},
  year={2017}
}

@article{ganea2018hyperbolic,
  title={Hyperbolic neural networks},
  author={Ganea, Octavian and B{\'e}cigneul, Gary and Hofmann, Thomas},
  journal={Advances in neural information processing systems},
  volume={31},
  year={2018}
}

@article{chami2019hyperbolic,
  title={Hyperbolic graph convolutional neural networks},
  author={Chami, Ines and Ying, Zhitao and R{\'e}, Christopher and Leskovec, Jure},
  journal={Advances in neural information processing systems},
  volume={32},
  year={2019}
}

@article{cohen2018spherical,
  title={Spherical cnns},
  author={Cohen, Taco S and Geiger, Mario and K{\"o}hler, Jonas and Welling, Max},
  journal={arXiv preprint arXiv:1801.10130},
  year={2018}
}

@inproceedings{esteves2018learning,
  title={Learning so (3) equivariant representations with spherical cnns},
  author={Esteves, Carlos and Allen-Blanchette, Christine and Makadia, Ameesh and Daniilidis, Kostas},
  booktitle={Proceedings of the european conference on computer vision (ECCV)},
  pages={52--68},
  year={2018}
}

@inproceedings{gu2019learning,
  title={Learning mixed-curvature representations in product spaces},
  author={Gu, Albert and Sala, Frederic and Gunel, Beliz and R{\'e}, Christopher},
  booktitle={International conference on learning representations},
  year={2018}
}

@article{skopek2020mixed,
  title={Mixed-curvature variational autoencoders},
  author={Skopek, Ondrej and Ganea, Octavian-Eugen and B{\'e}cigneul, Gary},
  journal={arXiv preprint arXiv:1911.08411},
  year={2019}
}

@article{brooks2019riemannian,
  title={Riemannian batch normalization for SPD neural networks},
  author={Brooks, Daniel and Schwander, Olivier and Barbaresco, Fr{\'e}d{\'e}ric and Schneider, Jean-Yves and Cord, Matthieu},
  journal={Advances in Neural Information Processing Systems},
  volume={32},
  year={2019}
}

@article{lou2020neural,
  title={Neural manifold ordinary differential equations},
  author={Lou, Aaron and Lim, Derek and Katsman, Isay and Huang, Leo and Jiang, Qingxuan and Lim, Ser Nam and De Sa, Christopher M},
  journal={Advances in Neural Information Processing Systems},
  volume={33},
  pages={17548--17558},
  year={2020}
}

@book{ungar2008gyrovector,
  title={A gyrovector space approach to hyperbolic geometry},
  author={Ungar, Abraham},
  year={2022},
  publisher={Springer Nature}
}

@inproceedings{ponce2016chalearn,
  title={Chalearn lap 2016: First round challenge on first impressions-dataset and results},
  author={Ponce-L{\'o}pez, V{\'\i}ctor and Chen, Baiyu and Oliu, Marc and Corneanu, Ciprian and Clap{\'e}s, Albert and Guyon, Isabelle and Bar{\'o}, Xavier and Escalante, Hugo Jair and Escalera, Sergio},
  booktitle={European conference on computer vision},
  pages={400--418},
  year={2016},
  organization={Springer}
}

@inproceedings{park2014pom,
author = {Park, Sunghyun and Shim, Han Suk and Chatterjee, Moitreya and Sagae, Kenji and Morency, Louis-Philippe},
title = {Computational Analysis of Persuasiveness in Social Multimedia: A Novel Dataset and Multimodal Prediction Approach},
year = {2014},
isbn = {9781450328852},
publisher = {Association for Computing Machinery},
address = {New York, NY, USA},
url = {https://doi.org/10.1145/2663204.2663260},
doi = {10.1145/2663204.2663260},
booktitle = {Proceedings of the 16th International Conference on Multimodal Interaction},
pages = {50–57},
numpages = {8},
location = {Istanbul, Turkey},
series = {ICMI '14}
}

@inproceedings{chen2016overcoming,
  title={Overcoming calibration problems in pattern labeling with pairwise ratings: application to personality traits},
  author={Chen, Baiyu and Escalera, Sergio and Guyon, Isabelle and Ponce-L{\'o}pez, V{\'\i}ctor and Shah, Nihar and Oliu Sim{\'o}n, Marc},
  booktitle={European Conference on Computer Vision},
  pages={419--432},
  year={2016},
  organization={Springer}
}

@article{shazeer2017outrageously,
  title={Outrageously large neural networks: The sparsely-gated mixture-of-experts layer},
  author={Shazeer, Noam and Mirhoseini, Azalia and Maziarz, Krzysztof and Davis, Andy and Le, Quoc and Hinton, Geoffrey and Dean, Jeff},
  journal={arXiv preprint arXiv:1701.06538},
  year={2017}
}

@article{fedus2022switch,
  title={Switch transformers: Scaling to trillion parameter models with simple and efficient sparsity},
  author={Fedus, William and Zoph, Barret and Shazeer, Noam},
  journal={Journal of Machine Learning Research},
  volume={23},
  number={120},
  pages={1--39},
  year={2022}
}

@article{mustafa2022multimodal,
  title={Multimodal contrastive learning with limoe: the language-image mixture of experts},
  author={Mustafa, Basil and Riquelme, Carlos and Puigcerver, Joan and Jenatton, Rodolphe and Houlsby, Neil},
  journal={Advances in Neural Information Processing Systems},
  volume={35},
  pages={9564--9576},
  year={2022}
}

@article{xin2025i2moe,
  title={I2MoE: Interpretable Multimodal Interaction-aware Mixture-of-Experts},
  author={Xin, Jiayi and Yun, Sukwon and Peng, Jie and Choi, Inyoung and Ballard, Jenna L and Chen, Tianlong and Long, Qi},
  journal={arXiv preprint arXiv:2505.19190},
  year={2025}
}

@article{he2021deberta,
  title={Debertav3: Improving deberta using electra-style pre-training with gradient-disentangled embedding sharing},
  author={He, Pengcheng and Gao, Jianfeng and Chen, Weizhu},
  journal={arXiv preprint arXiv:2111.09543},
  year={2021}
}

@article{baevski2020wav2vec,
  title={wav2vec 2.0: A framework for self-supervised learning of speech representations},
  author={Baevski, Alexei and Zhou, Yuhao and Mohamed, Abdelrahman and Auli, Michael},
  journal={Advances in neural information processing systems},
  volume={33},
  pages={12449--12460},
  year={2020}
}

@article{hsu2021hubert,
  title={Hubert: Self-supervised speech representation learning by masked prediction of hidden units},
  author={Hsu, Wei-Ning and Bolte, Benjamin and Tsai, Yao-Hung Hubert and Lakhotia, Kushal and Salakhutdinov, Ruslan and Mohamed, Abdelrahman},
  journal={IEEE/ACM transactions on audio, speech, and language processing},
  volume={29},
  pages={3451--3460},
  year={2021},
  publisher={IEEE}
}

@article{tong2022videomae,
  title={Videomae: Masked autoencoders are data-efficient learners for self-supervised video pre-training},
  author={Tong, Zhan and Song, Yibing and Wang, Jue and Wang, Limin},
  journal={Advances in neural information processing systems},
  volume={35},
  pages={10078--10093},
  year={2022}
}

@inproceedings{bertasius2021spacetime,
    author  = {Gedas Bertasius and Heng Wang and Lorenzo Torresani},
    title = {Is Space-Time Attention All You Need for Video Understanding?},
    booktitle   = {Proceedings of the International Conference on Machine Learning (ICML)}, 
    month = {July},
    year = {2021}
}

@article{loshchilov2018adamw,
  title={Decoupled weight decay regularization},
  author={Loshchilov, Ilya and Hutter, Frank},
  journal={arXiv preprint arXiv:1711.05101},
  year={2017}
}

@article{mccrae1992five,
  title={The five-factor model in personality: A critical appraisal},
  author={McAdams, Dan P},
  journal={Journal of personality},
  volume={60},
  number={2},
  pages={329--361},
  year={1992},
  publisher={Wiley Online Library}
}

@inproceedings{radford2023robust,
  title={Robust speech recognition via large-scale weak supervision},
  author={Radford, Alec and Kim, Jong Wook and Xu, Tao and Brockman, Greg and McLeavey, Christine and Sutskever, Ilya},
  booktitle={International conference on machine learning},
  pages={28492--28518},
  year={2023},
  organization={PMLR}
}

@article{thurstone1927law,
  title={{A law of comparative judgment}},
  author={Thurstone, Louis L.},
  journal={Psychological Review},
  volume={34},
  number={4},
  pages={273--286},
  year={1927},
  publisher={American Psychological Association (APA)}
}

@article{negahban2018learning,
  title={{Learning from Comparisons and Choices}},
  author={Negahban, Sahand and Oh, Sewoong and Thekumparampil, Kiran K. and Xu, Jiaming},
  journal={Journal of Machine Learning Research},
  volume={19},
  pages={1--95},
  year={2018}
}

@article{bradley1952rank,
  title={{Rank analysis of incomplete block designs. I. The method of paired comparisons}},
  author={Bradley, Ralph Allan and Terry, Milton E.},
  journal={Biometrika},
  volume={39},
  number={3/4},
  pages={324--345},
  year={1952},
  publisher={JSTOR}
}

@book{luce1959individual,
  title={{Individual choice behavior: A theoretical analysis}},
  author={Luce, R. Duncan},
  year={1959},
  publisher={Wiley}
}

@phdthesis{fazel2002matrix,
  title={{Matrix rank minimization with applications}},
  author={Fazel, Maryam},
  year={2002},
  school={Stanford University}
}

@article{cai2010singular,
  title={{A singular value thresholding algorithm for matrix completion}},
  author={Cai, Jian-Feng and Cand{\`e}s, Emmanuel J. and Shen, Zuowei},
  journal={SIAM Journal on Optimization},
  volume={20},
  number={4},
  pages={1956--1982},
  year={2010},
  publisher={SIAM}
}

@article{barzilai1988two,
  title={{Two-point step size gradient methods}},
  author={Barzilai, Jonathan and Borwein, Jonathan M.},
  journal={IMA journal of numerical analysis},
  volume={8},
  number={1},
  pages={141--148},
  year={1988},
  publisher={Oxford University Press}
}

@article{glickman1999rating,
  title={{Rating the chess rating system}},
  author={Glickman, Mark E. and Jones, Albyn C.},
  journal={Chance},
  volume={12},
  number={2},
  pages={21--28},
  year={1999},
  publisher={Taylor \& Francis}
}

@inproceedings{herbrich2007trueskill,
  title={{TrueSkill: A Bayesian Skill Rating System}},
  author={Herbrich, Ralf and Minka, Tom and Graepel, Thore},
  booktitle={Advances in Neural Information Processing Systems 19 (NIPS)},
  pages={569--576},
  year={2007}
}

@article{glickman1999parameter,
  title={{Parameter estimation in large dynamic paired comparison experiments}},
  author={Glickman, Mark E.},
  journal={Applied Statistics},
  volume={48},
  number={3},
  pages={377--394},
  year={1999},
  publisher={JSTOR}
}

@inproceedings{biel2011facetube,
  title        = {Facetube: Predicting Personality from Facial Expressions of Emotion in Online Conversational Video},
  author       = {Biel, Joan-Isaac and Gatica-Perez, Daniel},
  booktitle    = {Proceedings of the 13th International Conference on Multimodal Interaction (ICMI)},
  pages        = {53--56},
  year         = {2011},
  organization = {ACM}
}

@article{song2020interview2personality,
  title        = {Interview2Personality: A Multimodal Dataset for Predicting Personality Traits from Job Interviews},
  author       = {Song, Yuncheng and Yang, Fan and Huang, Sheng and Chen, Lin},
  journal      = {IEEE Transactions on Affective Computing},
  year         = {2020},
  doi          = {10.1109/TAFFC.2020.2978370}
}
}

\clearpage
\appendix

\section{\textsc{RecruitView} Dataset}
\label{app:dataset}

\subsection{Questions}
\label{app:questions}
The full list of the 76 unique interview questions used as prompts in the data collection is provided in Table~\ref{tab:questions}. These questions were curated from professional open-source resources, networking platforms, and consultations to elicit responses rich in both professional content and personality indicators.

\begin{table*}[!t] 
\centering
\begin{adjustbox}{width=\linewidth}
\begin{tabular}{rl | rl}
\toprule
\textbf{Sr. No.} & \textbf{Question} & \textbf{Sr. No.} & \textbf{Question} \\
\midrule
1. & Introduce yourself. & 41. & What are the three things that are most important for you in a job? \\
2. & What are your greatest strengths and weaknesses? & 42. & How did you handle disagreements? \\
3. & How do you handle changes or unexpected situations in the workplace? & 43. & Tell me about a time where you experienced difficulty while working on a project. How did you handle it? \\
4. & What is your biggest achievement so far? & 44. & What makes you happy? \\
5. & Tell me about a time when you went above and beyond the call of duty to achieve a goal or deliver results. & 45. & Can you give an example of a situation where you mentored a junior colleague, helping them grow professionally and personally? \\
6. & Give me an example of your creativity. & 46. & What are you passionate about? \\
7. & How do you work under pressure? Can you handle the pressure? & 47. & What motivates you to perform at your best in the workplace? \\
8. & If you won a Rs.10-crore lottery, would you still work? & 48. & Describe a time when you proactively sought out opportunities to develop new skills or knowledge relevant to your role. \\
9. & What motivates you? & 49. & Can you give an example of a situation where you leveraged technology or automation to streamline a process and increase efficiency? \\
10. & Can you give an example of a time when you successfully implemented a solution to improve a process or procedure? & 50. & Share a story of a project where you collaborated with a cross-functional team to deliver exceptional results. \\
11. & Tell me about a time when you had to step into a role outside of your expertise to support the team's objectives. & 51. & Describe a situation where you had to adapt to a change in the work environment. \\
12. & How do you respond to change? & 52. & What are you most proud of? \\
13. & What was the toughest decision you ever had to make? & 53. & What do you think is an ideal work environment? \\
14. & What is your greatest fear? & 54. & Tell me about a time you initiated or led that had a positive impact on your team or organization. \\
15. & Describe a project where you took the lead in implementing a new strategy or process, driving positive change within your team or organization. & 55. & Can you give an example of a time when you had to resolve a disagreement or misunderstanding within a team? \\
16. & Describe a situation where you identified a problem before it became significant. What steps did you take to address it? & 56. & How do you deal with criticism? \\
17. & How would you rate yourself on a scale of 1 to 10? & 57. & Tell me about a time when you failed to meet a goal or objective. How did you handle it? \\
18. & How do you handle stress and anxiety? & 58. & What has been your greatest failure? \\
19. & Tell me about a time when you were not satisfied with your performance. & 59. & Tell me about a time when you had to resolve a conflict with a coworker or team member. \\
20. & Can you give an example of a time when you successfully managed multiple competing priorities? & 60. & Share a story of a project where you led the team in developing and implementing a solution that resulted in significant cost savings or revenue growth. \\
21. & Where do you see yourself in the next 5 years? & 61. & Tell me about a time when you had to work on a project outside of your comfort zone. How did you handle it? \\
22. & Why should a company hire you? & 62. & Are you an organized person? \\
23. & Are you reliable or can I trust you with responsibilities? & 63. & What do you always regret, or do you have any regrets? \\
24. & What makes you angry? & 64. & Share a story of how you took ownership of a project that was struggling and turned it into a success through your initiative. \\
25. & Can you give an example of a time when you had to persuade others to adopt your ideas or proposals? & 65. & Can you give an example of a situation where you successfully motivated a disengaged team member to contribute effectively to a project? \\
26. & Can you give an example of a time when you had to take the initiative to solve a problem without being asked? & 66. & How do you learn new skills? \\
27. & Are you open to taking risks or do you like experimenting? & 67. & Describe a situation where you had to prioritize tasks under tight deadlines. \\
28. & What is your dream company like? & 68. & How quickly do you adapt to new technology? \\
29. & Do you have a good work ethic? & 69. & Can you give an example of a time when you facilitated a productive team meeting or discussion? \\
30. & Tell me about a time when you recognized and capitalized on the unique strengths of individual team members to achieve a common goal. & 70. & Describe a project where you collaborated with stakeholders to define the problem and develop a solution that met everyone's needs. \\
31. & Tell me about a time when you successfully resolved a long-standing issue that had been impeding progress within your team or organization. & 71. & What is your dream job like? \\
32. & How do you improve your knowledge? & 72. & What are your weaknesses? \\
33. & What are your hobbies? & 73. & Tell me about a time when you proposed an innovative idea that significantly improved team efficiency or productivity. \\
34. & Can you give an example of a time when you led by example to promote a positive work culture or values? & 74. & Can you give an example of a time when you coached or mentored a colleague to help them achieve their goals? \\
35. & Describe a project where you encouraged open communication and feedback among team members, leading to improved collaboration and outcomes. & 75. & Share a story of a time when you rallied your team during a crisis, fostering resilience and determination. \\
36. & Is there anything that makes you different from other candidates? & 76. & Tell me about yourself. \\
37. & Can you describe your time management skills? & & \\
38. & Can you describe a situation where you had to overcome a significant challenge in a team setting? & & \\
\bottomrule
\end{tabular}
\end{adjustbox}
\caption{The 76 curated interview questions used as prompts in the \textsc{RecruitView} dataset.}
\label{tab:questions}
\end{table*} 

\subsection{Collection and Labeling Framework}
\label{app:websites}

\begin{figure*}[t]
    \centering
    \includegraphics[width=\linewidth]{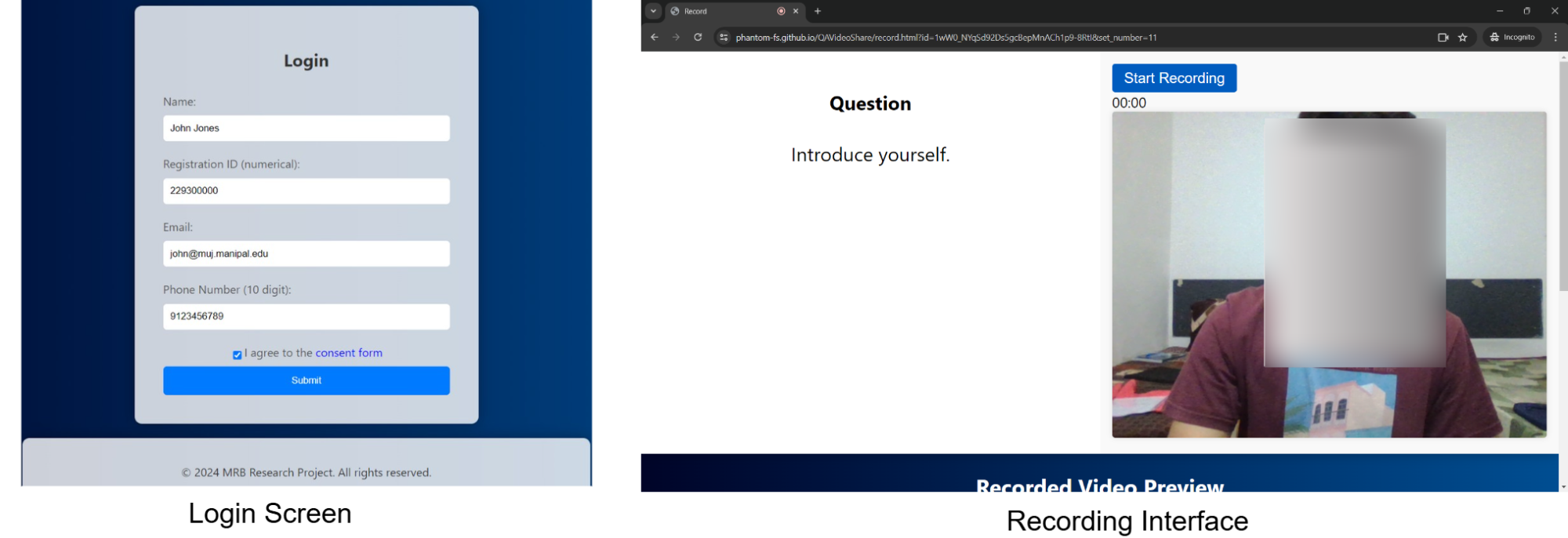} 
    \caption{The participant-facing \texttt{QAVideoShare} data collection platform. (Left) The secure login and consent portal. (Right) The primary video recording interface where participants view the prompt and record their response.}
    \label{fig:interview_platform}
\end{figure*}

To ensure standardized data acquisition and annotation, we developed two custom web-based platforms. The first, \texttt{QAVideoShare}\footnote{\url{https://github.com/Phantom-fs/QAVideoShare}}, is an online interview platform designed to collect video responses. Participants were presented with questions and recorded unscripted answers directly through the browser interface, ensuring uniform question presentation and automated video storage. The participant's workflow, from authentication to recording, is shown in Figure \ref{fig:interview_platform}.

\begin{figure*}[t]
    \centering
    \includegraphics[width=\linewidth]{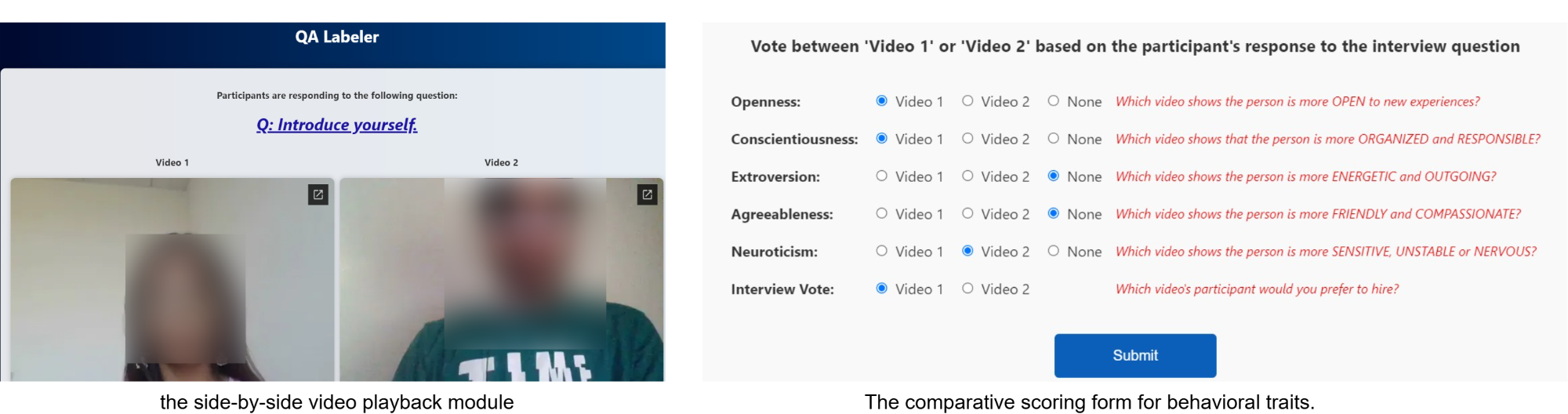} 
    \caption{The evaluator-facing \texttt{QA-Labeler} annotation platform. (Left) The side-by-side video playback module for comparative judgment. (Right) The corresponding scoring form where evaluators provide choice ratings on behavioral traits.}
    \label{fig:labeling_platform}
\end{figure*}

The second platform, \texttt{QA-Labeler}\footnote{\url{https://github.com/Phantom-fs/QA-Labeler}}, was developed for data labeling and evaluation. This tool allowed clinical psychologists (annotators) to view recorded videos and provide comparative assessments across various behavioral and performance criteria. The comparative judgment interface, featuring a side-by-side player and scoring form, is detailed in Figure \ref{fig:labeling_platform}. Both platforms support browser-based, multi-user operation, enabling a scalable and consistent data processing pipeline.

\subsection{Labeling Comparison}
\label{app:labeling_comparison}
To ensure that the conversion of pairwise judgments into continuous rankings was both consistent and interpretable, we evaluated five independent label-conversion frameworks: \textbf{Glicko-2}, \textbf{TrueSkill}, \textbf{Full-Rank MNL}, \textbf{MNL-with-Ties}, and the \textbf{Nuclear-Norm-Regularized MNL}. Each method produced a scalar ranking score representing the latent position of each video across all pairwise comparisons. All models were trained on identical data.

\subsubsection{Ground-Truth Construction}
We adopt a leave-one-out consensus evaluation. When evaluating a given model, its output is compared against a ground-truth defined as the mean of the standardized (Z-scored) scores from all \emph{other} models. This avoids self-evaluation, ensures symmetric treatment of all frameworks, and controls for scale or range mismatches. For models such as Nuclear-Norm MNL which inherently output normalized scores, further standardization was not applied.

\subsubsection{Evaluation Results}
\begin{figure}[!t]
\centering
\includegraphics[width=\linewidth]{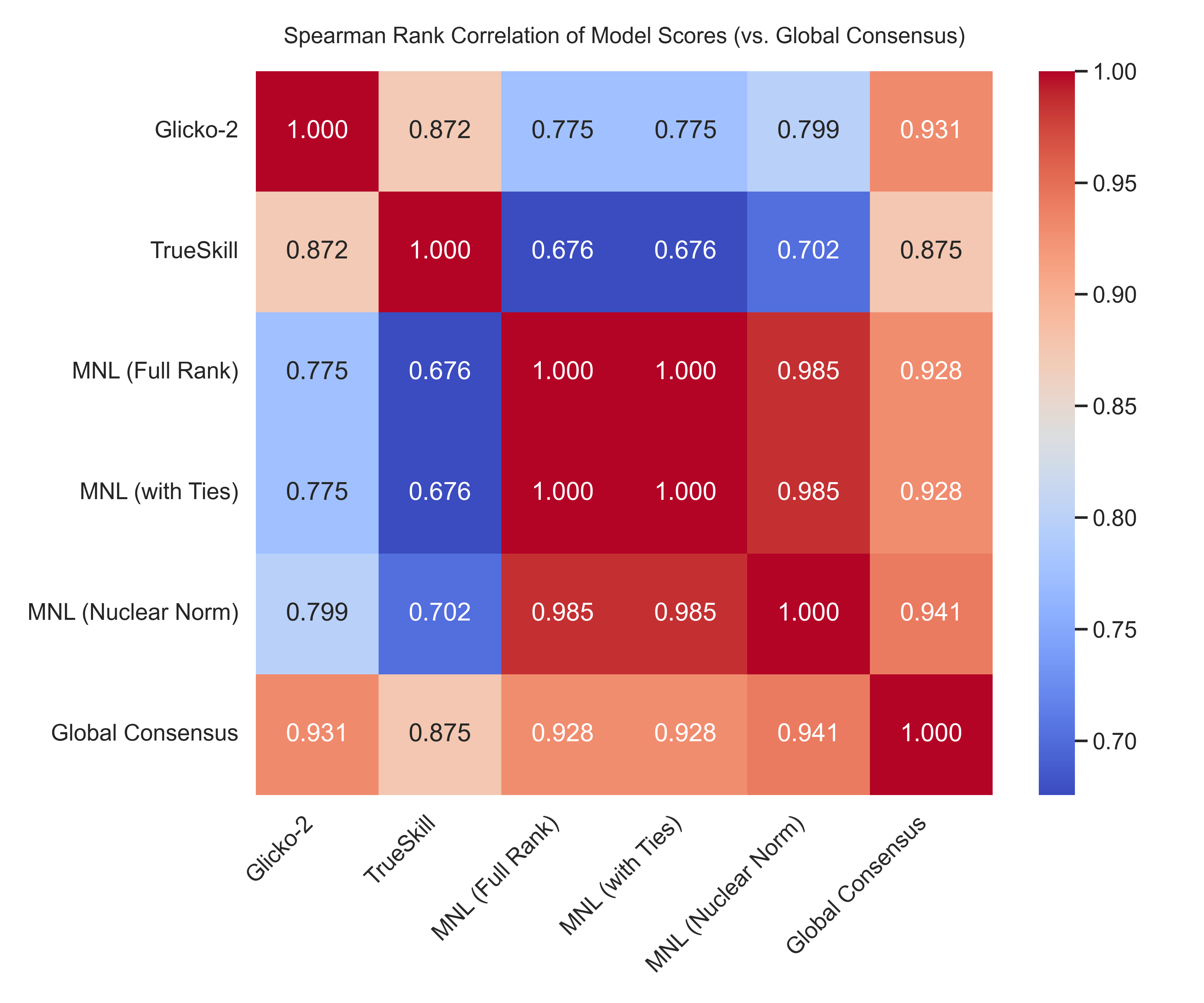}
\caption{Correlation matrix among the five label-conversion frameworks.}
\label{fig:corr_heatmap}
\end{figure}

\begin{figure}[!t]
\centering
\includegraphics[width=\linewidth]{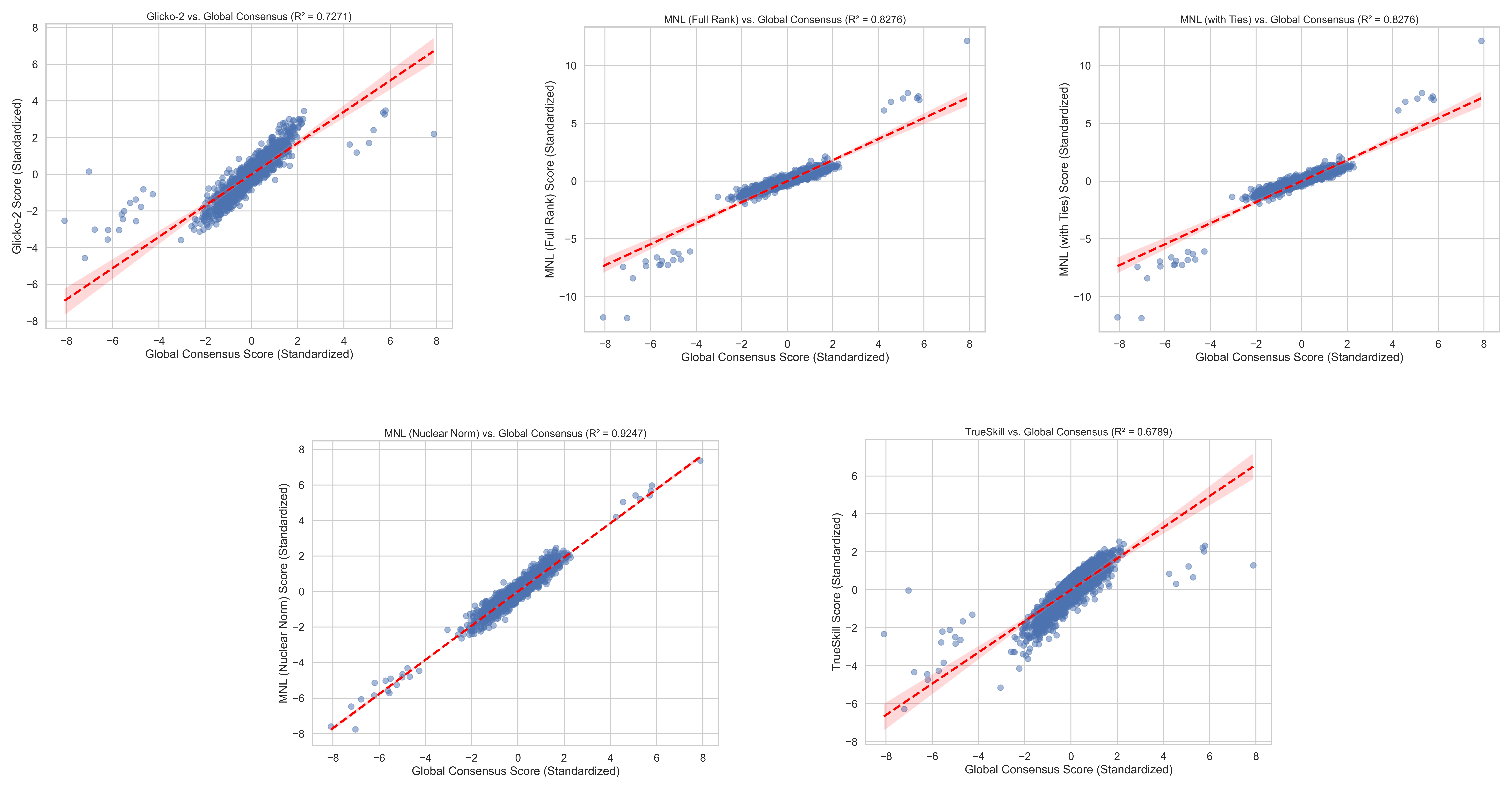}
\caption{Scatter plots of predicted vs. ground-truth rankings for all five label-conversion frameworks. Each subplot corresponds to one model.}
\label{fig:scatter_combined}
\end{figure}

Evaluation was performed using \textbf{Spearman's~$\rho$}, \textbf{Kendall's~$\tau$}, \textbf{RMSE}, \textbf{MAE}, and \textbf{Precision/Recall@10\%}, computed on continuous ranking outputs. These metrics were chosen for their compatibility with ordinal data, as our objective is to evaluate \emph{relative ordering} fidelity rather than categorical correctness. Figure~\ref{fig:corr_heatmap} shows the Spearman rank-correlation structure across models. Here, higher correlation is desirable because label-conversion methods are not supposed to invent disagreement; divergence between methods would indicate instability or method-specific distortion rather than genuine latent behavioral differences. Figure~\ref{fig:scatter_combined} further reinforces this observation, showing all five models plotted against the consensus reference for direct visual comparison.

Although all frameworks produced broadly compatible rankings, \textbf{Full-Rank MNL} achieved slightly higher peak correlations on isolated traits, while the \textbf{Nuclear-Norm MNL} exhibited greater overall stability with low variance across random drop-model trials ($\rho = 0.905 \pm 0.04$). The low-rank constraint enforces smoother coupling across correlated personality traits, yielding more stable global rankings; this behavior is further reflected in the robustness summary (Table~\ref{tab:robustness_check}).

\subsubsection{Secondary Verification and Qualitative Assessment}
A secondary verification was conducted through a ratio-based test: a manually selected subset of pairwise comparisons was converted into empirical win–loss ratios, which serve as a local ordinal reference. The Nuclear-Norm MNL produced the closest match, accurately preserving both relative order and proportional differences. A small leaderboard test confirmed that local chains (A $>$ B $>$ C) remained globally consistent (A $>$ C) and aligned with human expectations. In qualitative inspection, videos ranked higher by this model displayed clearer articulation, stronger confidence, and more natural expressiveness.

Accordingly, we adopt the Nuclear-Norm formulation as the final label-conversion framework for \textsc{RecruitView}. Its low-rank structure offered smoother scaling across correlated targets, and its predictions were the most consistent with manual verification.

\begin{table}[!t]
\centering
\begin{tabular}{lccc}
\toprule
\textbf{Model} & \textbf{Avg. $\rho$} & \textbf{Std. Dev.} & \textbf{Trials} \\
\midrule
MNL (Full Rank) & 0.910 & 0.044 & 13 \\
MNL (with Ties) & 0.903 & 0.042 & 17 \\
MNL (Nuclear Norm) & {0.905} & {0.040} & 19 \\
Glicko-2 & 0.860 & 0.049 & 16 \\
TrueSkill & 0.776 & 0.053 & 15 \\
\bottomrule
\end{tabular}
\caption{Robustness check across 20 random drop-model trials.}
\label{tab:robustness_check}
\end{table}

\subsection{Detailed Data Statistics}
\label{app:data_stats}

Table~\ref{tab:feature_stats} presents comprehensive summary statistics for the 2,011 video segments in \textsc{RecruitView}. The clips have a mean duration of 29.66 seconds ($\sigma=16.40$), with a minimum of 0.60 seconds and a maximum of 92.34 seconds. This temporal range ensures models are exposed to both brief ``thin-slice" judgments and longer-form analyses. The transcripts are similarly diverse, with a mean word count of 81.90 ($\sigma=51.15$) and a maximum of 266 words, providing a rich linguistic substrate for multimodal analysis. The median (50th percentile) values for duration (27.27s) and word count (72.00) closely track their respective means, confirming the well-behaved nature of these distributions.

\begin{table}[!t]
\centering
\begin{tabular}{lcc}
\toprule
\textbf{Statistic} & \textbf{Duration (seconds)} & \textbf{Word Count} \\
\midrule
count & 2011.00 & 2011.00 \\
mean & 29.66 & 81.90 \\
std & 16.40 & 51.15 \\
min & 0.60 & 0.00 \\
25\% & 15.83 & 41.00 \\
50\% & 27.27 & 72.00 \\
75\% & 42.63 & 115.00 \\
max & 92.34 & 256.00 \\ 
\bottomrule
\end{tabular}
\caption{Statistics for Engineered Features in \textsc{RecruitView}. The table shows distribution statistics for video duration (in seconds) and transcript word count across all 2,011 clips.}
\label{tab:feature_stats}
\end{table}

\subsection{Complete Correlation Matrix}
\label{subsec:full_correlation}

\begin{figure}[!t]
    \centering
    \includegraphics[width=\linewidth]{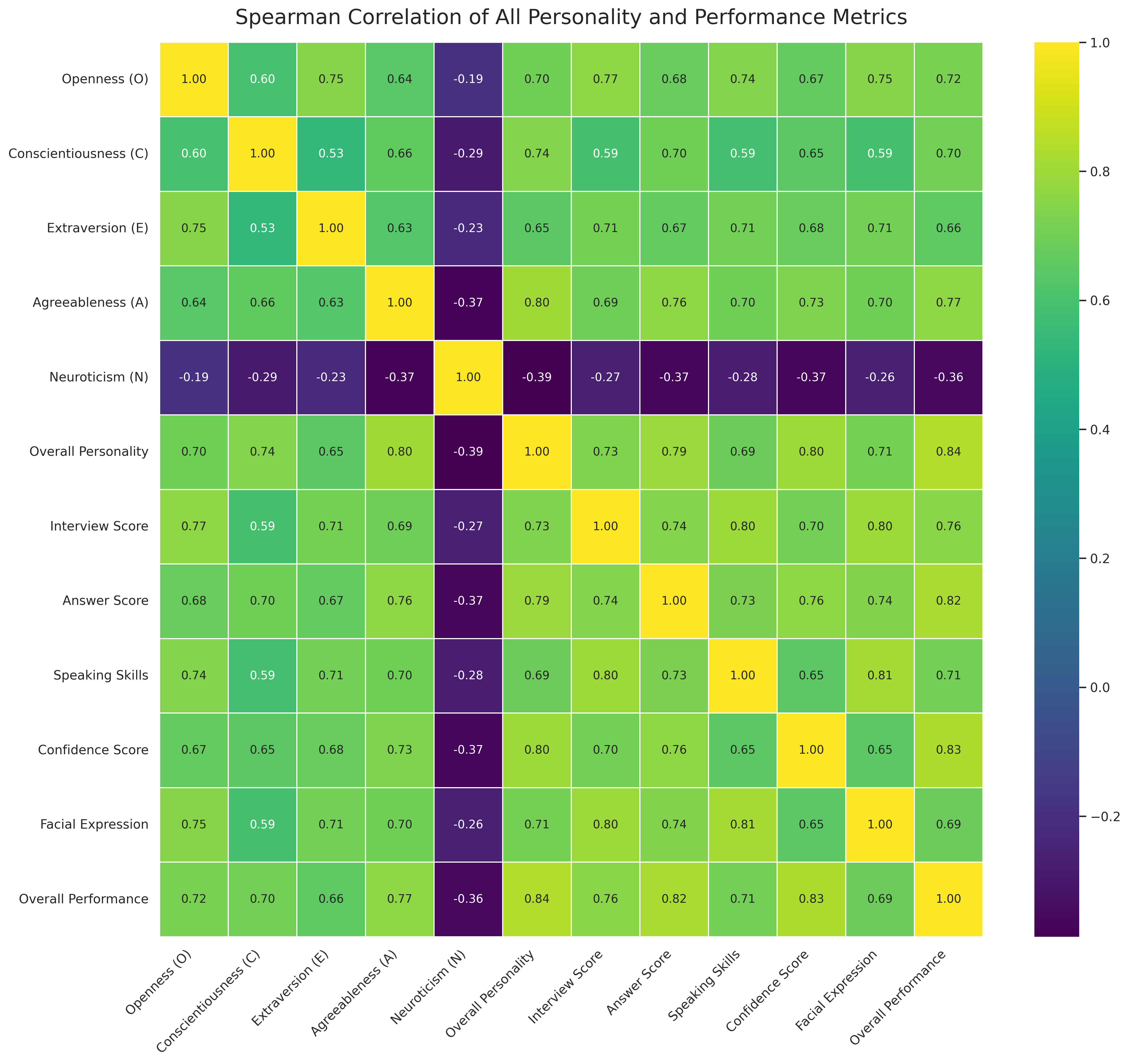}
    \caption{Spearman’s $\rho$ correlation matrix for all 12 metrics.}
    \label{fig:corr_cross_domain}
\end{figure}

Table~\ref{tab:correlation_matrix} presents the complete Spearman correlation matrix across all 12 target dimensions in \textsc{RecruitView}. This comprehensive view consolidates the patterns observed in Figure~\ref{fig:corr_cross_domain}, revealing the full structure of dependencies among the Big Five personality traits (O, C, E, A, N), Overall Personality, and the six interview performance metrics (Interview Score, Answer Score, Speaking Skills, Confidence Score, Facial Expression, and Overall Performance). The matrix exhibits several key characteristics: strong positive correlations within the personality cluster (upper-left block) and performance cluster (bottom-right block), moderate positive correlations in the cross-domain blocks, and consistent negative correlations involving Neuroticism across all dimensions. The cross-correlation block (bottom-left) shows intuitive patterns:
\begin{itemize}
    \item \textit{Extraversion} is positively correlated with \textit{Speaking Skills} ($\rho = 0.71$) and \textit{Facial Expression} ($\rho = 0.71$), suggesting that outgoing individuals are perceived as more expressive and articulate.
    \item \textit{Conscientiousness} shows a clear positive relationship with \textit{Answer Score} ($\rho = 0.70$), aligning with the expectation that diligent individuals provide higher-quality responses.
    \item \textit{Neuroticism} demonstrates a consistent negative correlation across all performance metrics, most notably with \textit{Confidence Score} ($\rho = -0.37$) and \textit{Overall Performance} ($\rho = -0.36$).
\end{itemize}

These structured dependencies highlight the interconnected nature of personality perception and observable interview behaviors, providing insights into which combinations of traits and performance indicators are most strongly linked in evaluative contexts.

\begin{table*}[!t]
\centering
\begin{adjustbox}{width=\textwidth}
\begin{tabular}{l|cccccc|cccccc}
\toprule
\textbf{Metrics} & \textbf{O} & \textbf{C} & \textbf{E} & \textbf{A} & \textbf{N} & \textbf{Pers.} & \textbf{Int.} & \textbf{Ans.} & \textbf{Spk.} & \textbf{Conf.} & \textbf{Fac.} & \textbf{Perf.} \\
\midrule
Openness (\textbf{O}) & 1.00 & 0.60 & 0.75 & 0.64 & -0.19 & 0.70 & 0.77 & 0.68 & 0.74 & 0.67 & 0.75 & 0.72 \\
Conscientiousness (\textbf{C}) & 0.60 & 1.00 & 0.53 & 0.66 & -0.29 & 0.74 & 0.59 & 0.70 & 0.59 & 0.65 & 0.59 & 0.70 \\
Extraversion (\textbf{E}) & 0.75 & 0.53 & 1.00 & 0.63 & -0.23 & 0.65 & 0.71 & 0.67 & 0.71 & 0.68 & 0.71 & 0.66 \\
Agreeableness (\textbf{A}) & 0.64 & 0.66 & 0.63 & 1.00 & -0.37 & 0.80 & 0.69 & 0.78 & 0.70 & 0.73 & 0.70 & 0.77 \\
Neuroticism (\textbf{N}) & -0.19 & -0.29 & -0.23 & -0.37 & 1.00 & -0.39 & -0.27 & -0.37 & -0.28 & -0.37 & -0.26 & -0.36 \\
Overall Personality (\textbf{Pers.}) & 0.70 & 0.74 & 0.65 & 0.80 & -0.39 & 1.00 & 0.73 & 0.79 & 0.69 & 0.80 & 0.71 & 0.84 \\
\midrule
Interview Score (\textbf{Int.}) & 0.77 & 0.59 & 0.71 & 0.69 & -0.27 & 0.73 & 1.00 & 0.74 & 0.80 & 0.70 & 0.80 & 0.76 \\
Answer Score (\textbf{Ans.}) & 0.68 & 0.70 & 0.67 & 0.78 & -0.37 & 0.79 & 0.74 & 1.00 & 0.73 & 0.76 & 0.74 & 0.82 \\
Speaking Skills (\textbf{Spk.}) & 0.74 & 0.59 & 0.71 & 0.70 & -0.28 & 0.69 & 0.80 & 0.73 & 1.00 & 0.65 & 0.81 & 0.71 \\
Confidence Score (\textbf{Conf.}) & 0.67 & 0.65 & 0.68 & 0.73 & -0.37 & 0.80 & 0.70 & 0.76 & 0.65 & 1.00 & 0.65 & 0.83 \\
Facial Expression (\textbf{Fac.}) & 0.75 & 0.59 & 0.71 & 0.70 & -0.26 & 0.71 & 0.80 & 0.74 & 0.81 & 0.65 & 1.00 & 0.69 \\
Overall Performance (\textbf{Perf.}) & 0.72 & 0.70 & 0.66 & 0.77 & -0.36 & 0.84 & 0.76 & 0.82 & 0.71 & 0.83 & 0.69 & 1.00 \\
\bottomrule
\end{tabular}
\end{adjustbox}
\caption{Complete Spearman Correlation Matrix for all 12 metrics in \textsc{RecruitView}.}
\label{tab:correlation_matrix}
\end{table*}

\subsection{Metrics Statistics}
\label{app:metrics_stats_analysis}

\begin{figure*}[!t]
    \centering
    \includegraphics[width=\linewidth]{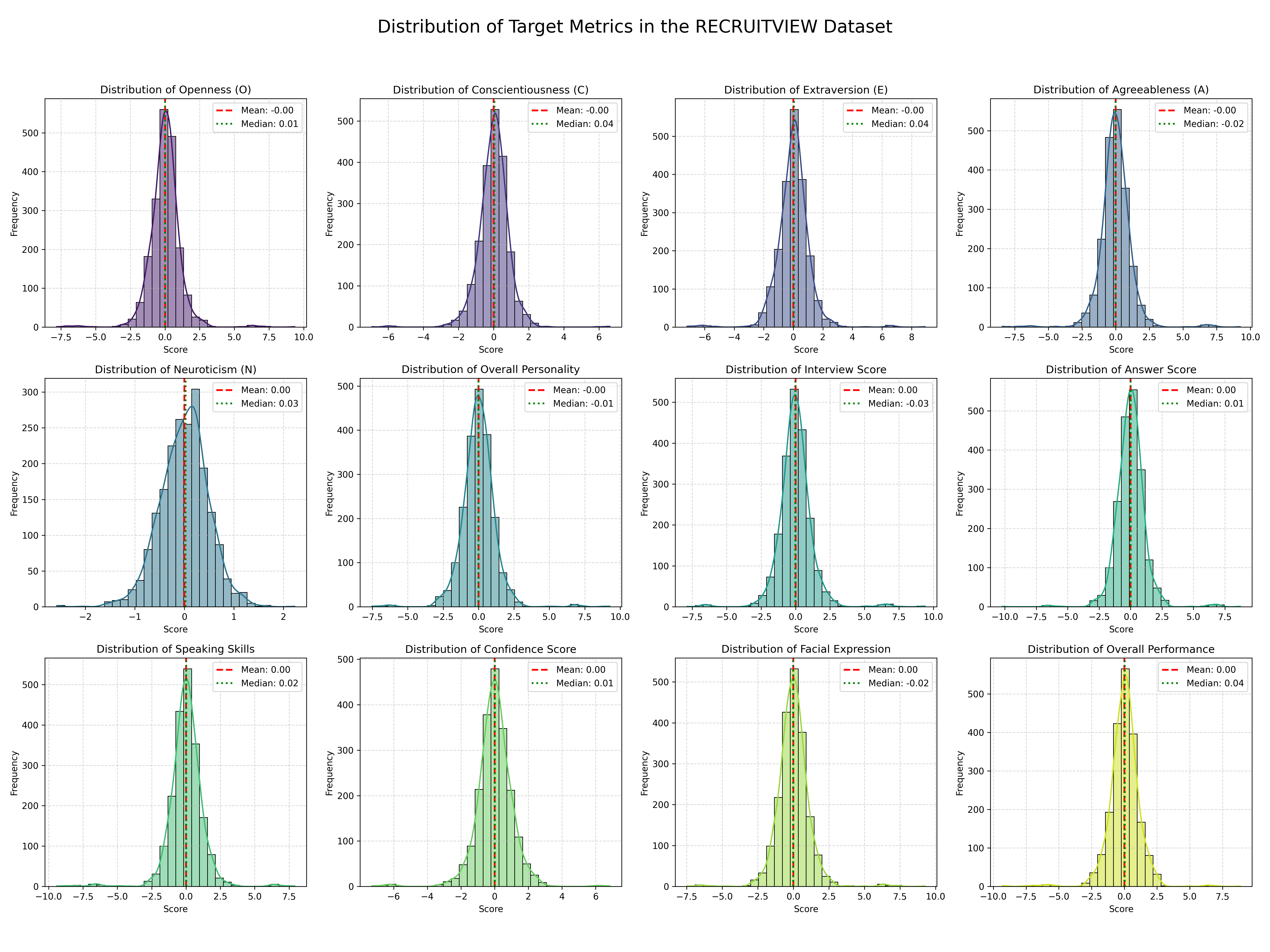}
    \caption{Distribution histograms for all 12 target metrics in \textsc{RecruitView}. Each metric is normalized with a mean near zero. The plots show varying degrees of skewness and heavy tails (leptokurtosis), motivating the use of robust loss functions and rank-based evaluation.}
    \label{fig:metrics_distributions}
\end{figure*}

Table~\ref{tab:metrics_statistics} and Figure~\ref{fig:metrics_distributions} jointly characterize the empirical behavior of all twelve targets in \textsc{RecruitView}.
First, the \emph{means} sit essentially at zero for every dimension (see ``mean'' row), confirming that the normalization pipeline yields centered targets and ensuring that absolute intercepts are not informative. The \emph{medians} closely track the means (50\% row in Table~\ref{tab:metrics_statistics}), and the modal mass of each histogram is concentrated around the origin (Figure~\ref{fig:metrics_distributions}), indicating a near-symmetric \emph{core} for most variables.

\noindent\textbf{Dispersion and dynamic range.}
Standard deviations cluster in the $[0.88, 1.28]$ interval for 11/12 metrics, with \textbf{Neuroticism (N)} exhibiting a much tighter spread ($\mathrm{std}=0.49$). This implies that N is intrinsically less variable across our population relative to other psychological or performance attributes. Conversely, Interview-adjacent outcomes (Int., Ans., Spk., Conf., Fac., Perf.) show broadly comparable dispersion ($\approx 1.1$–$1.28$), desirable for multi-task optimization with shared heads. The \emph{extrema} reveal long tails for several metrics (e.g., Ans.: $\min=-10.20$; O: $\max=9.34$), which are far beyond $\pm 3\sigma$ and thus constitute influential observations for any squared-loss estimator.

\noindent\textbf{Asymmetry (skewness).}
Skewness in Table~\ref{tab:metrics_statistics} uncovers systematic asymmetries:
\begin{itemize}\setlength{\itemsep}{2pt}
    \item \emph{Negative skew} for \textbf{C}, \textbf{Spk}, \textbf{Conf}, and \textbf{Perf} ($-0.57$, $-0.86$, $-0.64$, $-0.75$) indicates heavier left tails and a right-shifted bulk. Practically, a larger fraction of samples achieve above-average performance on speaking, confidence, and overall performance, with relatively fewer but more extreme low outliers.
    \item \emph{Positive skew} for \textbf{A}, \textbf{Pers.}, \textbf{Int.}, and \textbf{Fac.} ($0.40$–$0.66$) suggests the opposite: mass slightly left of zero with occasional high outliers. \textbf{Openness} and \textbf{Extraversion} show mild asymmetry ($0.03$ and $-0.22$), while \textbf{Neuroticism} is modestly negative ($-0.25$), again consistent with its compressed variance.
\end{itemize}
These asymmetries imply that symmetric error models may under- or over-penalize different tails across tasks; model selection should therefore consider robust losses and rank-based metrics.

\noindent\textbf{Tail heaviness (kurtosis).}
All targets except \textbf{Neuroticism} show pronounced leptokurtosis (excess kurtosis $\approx 8.8$–$13.4$), confirming heavy tails and a high concentration near the center (Table~\ref{tab:metrics_statistics}). \textbf{Neuroticism} ($1.14$) is notably closer to mesokurtic behavior relative to the other metrics. Combined with the extreme minima/maxima, this indicates that a small subset of clips carry disproportionately informative deviations—a regime where (i) Huber/quantile losses and (ii) clipping or winsorization, materially improve stability and interpretability.

\noindent\textbf{Quartiles and central mass.}
Interquartile ranges are tightly packed around zero (25\%–75\% roughly $\pm 0.55$–$0.62$ for most metrics), reinforcing that the majority of ratings occupy a narrow band. The practical upshot is twofold: (a) small absolute errors around the origin correspond to meaningful rank changes, and (b) evaluation should prioritize \emph{monotonicity} (\emph{Spearman} $\rho$, \emph{Kendall} $\tau$, or concordance index) in addition to pointwise deviations.

\begin{table*}[!t]
\centering
\scriptsize
\setlength{\tabcolsep}{3pt}
\renewcommand{\arraystretch}{1.1}
\begin{adjustbox}{width=\linewidth}
\begin{tabular}{l*{12}{r}}
\toprule
\textbf{Statistic} & \textbf{O} & \textbf{C} & \textbf{E} & \textbf{A} & \textbf{N} & \textbf{Pers.} & \textbf{Int.} & \textbf{Ans.} & \textbf{Spk.} & \textbf{Conf.} & \textbf{Fac.} & \textbf{Perf.} \\
\midrule
count & 2011.000 & 2011.000 & 2011.000 & 2011.000 & 2011.000 & 2011.000 & 2011.000 & 2011.000 & 2011.000 & 2011.000 & 2011.000 & 2011.000 \\
mean & -0.000 & -0.000 & -0.000 & -0.000 & -0.000 & -0.000 & 0.000 & 0.000 & 0.000 & 0.000 & 0.000 & 0.000 \\
std & 1.127 & 0.877 & 1.098 & 1.200 & 0.487 & 1.203 & 1.230 & 1.176 & 1.278 & 1.080 & 1.151 & 1.201 \\
min & -7.080 & -6.929 & -7.184 & -8.412 & -2.577 & -7.526 & -7.860 & -10.199 & -9.406 & -7.286 & -7.461 & -9.311 \\
25\% & -0.540 & -0.448 & -0.546 & -0.566 & -0.306 & -0.611 & -0.596 & -0.611 & -0.558 & -0.535 & -0.584 & -0.575 \\
50\% & 0.009 & 0.043 & 0.044 & -0.016 & 0.026 & -0.011 & -0.027 & 0.006 & 0.020 & 0.011 & -0.018 & 0.043 \\
75\% & 0.522 & 0.485 & 0.544 & 0.548 & 0.303 & 0.617 & 0.545 & 0.579 & 0.616 & 0.553 & 0.549 & 0.601 \\
max & 9.339 & 6.617 & 8.910 & 9.242 & 2.226 & 9.270 & 9.386 & 8.722 & 7.901 & 6.839 & 9.256 & 8.049 \\
mode & -7.080 & -6.929 & -7.184 & -8.412 & -2.577 & -7.526 & -7.060 & -10.199 & -9.406 & -7.286 & -7.461 & -9.311 \\
skew & 0.027 & -0.570 & -0.218 & 0.398 & -0.245 & 0.462 & 0.660 & 0.353 & -0.855 & -0.640 & 0.598 & -0.748 \\
kurtosis & 12.408 & 10.763 & 11.960 & 13.445 & 1.140 & 10.499 & 11.604 & 12.033 & 12.727 & 8.821 & 11.179 & 11.742 \\
\bottomrule
\end{tabular}
\end{adjustbox}
\caption{Comprehensive statistical summary of all 12 target dimensions in \textsc{RecruitView}. The table shows distribution statistics including central tendency, dispersion, range, and shape measures for the Big Five personality traits (O=Openness, C=Conscientiousness, E=Extraversion, A=Agreeableness, N=Neuroticism), Overall Personality (Pers.), and six interview performance metrics (Int.=Interview Score, Ans.=Answer Score, Spk.=Speaking Skills, Conf.=Confidence Score, Fac.=Facial Expression, Perf.=Overall Performance). Near-zero means confirm proper normalization, while the skewness and kurtosis values indicate the presence of outliers and heavy tails in some dimensions.}
\label{tab:metrics_statistics}
\end{table*}

\subsection{Data Splits}
We use stratified random sampling to create training (70\%, 1404 samples), validation (15\%, 290 samples), and test (15\%, 317 samples) splits. Stratification is performed on the anonymized user number (i.e., ID) to prevent identity leakage across data splits. The same splits are used for all experiments to enable fair comparison.


\subsection{Metadata}
\label{app:json_sample}

Each entry in the \textsc{RecruitView} metadata file follows the structure shown below. Note that personally identifiable information (user name) has been anonymized.

\begin{lstlisting}[language=json, basicstyle=\tiny\ttfamily, breaklines=true]
{
    "id": "0001",
    "video_id": "vid_0001",
    "video_filename": "vid_0001.mp4",
    "duration": "long",
    "question_id": "1",
    "question": "Introduce yourself",
    "video_quality": "High",
    "user_no": "147",
    "Openness (O)": -0.653,
    "Conscientiousness (C)": -0.049,
    "Extraversion (E)": -0.691,
    "Agreeableness (A)": -0.293,
    "Neuroticism (N)": 0.190,
    "overall_personality": -0.029,
    "interview_score": -0.923,
    "answer_score": -0.803,
    "speaking_skills": -0.769,
    "confidence_score": -0.362,
    "facial_expression": -0.817,
    "overall_performance": -0.456,
    "transcript": "[00:01 - 00:11] Hello everyone, this is ..."
}
\end{lstlisting}

The twelve continuous scores are normalized and represent relative performance across the dataset, derived from the nuclear-norm regularized MNL model described in Section~\ref{subsec:data_format}.

\section{Ethics}
\label{app:ethics}

\subsection{Participant Protection and Data Collection}
\label{sec:ethics_participants}

All data collection procedures for the \textsc{RecruitView} dataset were conducted under institutional ethical approval and followed human research standards consistent with the Declaration of Helsinki. Participants were fully briefed about the study’s purpose and provided written informed consent prior to participation. The consent form explicitly covered the recording of interview videos, data usage for academic research, and the voluntary nature of participation. Participants were informed of their right to withdraw their data at any point before public release of the dataset used in this study, without consequence. No personally identifiable information (PII) was stored alongside the recordings. All metadata were anonymized, and the participant entries are linked only to an anonymized user number.

Participants were recruited through voluntary university outreach programs and online calls for participation. The pool consisted primarily of adult volunteers, with no inclusion of vulnerable populations.

\subsection{Data Annotation and Labeling}
\label{sec:ethics_annotation}

Annotations were performed by clinical psychologists familiar with behavioral and personality assessment protocols. A pairwise comparison framework was adopted instead of absolute rating to reduce inter-rater calibration bias and to ensure consistency across annotators. Comparative judgments were aggregated using a nuclear-norm regularized multinomial logit model to derive continuous, psychometrically consistent target scores. Annotators were compensated fairly for their professional effort. All annotator identities remain confidential.

\subsection{Dataset and Model: Bias, Misuse, and Fairness}
\label{sec:ethics_bias}

The dataset’s participant pool exhibits diversity in gender, accent, and educational background, but full demographic uniformity is not available. As such, models trained on this dataset may not generalize equally across other population subgroups. We explicitly acknowledge this limitation and encourage future fairness audits. The \textsc{RecruitView} dataset and the associated \texttt{CRMF} model are intended solely for research on multimodal behavioral and personality assessment. They are not validated for real-world deployment, hiring processes, or psychological diagnostics. Any attempt to use this work for employment screening, psychological profiling, or commercial analytics constitutes misuse. Although care was taken to minimize annotation bias and maintain fairness, all data-driven systems may still inherit spurious correlations; future work will include comprehensive fairness and subgroup analyses.

\subsection{Responsible Research Practices}
\label{sec:ethics_responsible}

We emphasize transparency regarding dataset scope and limitations, including moderate dataset size and short interview durations. The dataset and code are released for non-commercial academic research to enable independent verification, benchmarking, and fairness assessment by the broader community.

Access to the \textsc{RecruitView} dataset is managed through a secure request portal. Applicants must submit an access request and sign a data usage agreement confirming (1) exclusive non-commercial academic use, (2) adherence to participant anonymity, and (3) compliance with the ethical guidelines outlined in this paper. The agreement prohibits using the dataset or models for employment decisions, identity profiling, or any commercial product development. Each access request is manually reviewed, and credentials are issued only upon approval and formal consent acknowledgment. Access logs are maintained, and the authors reserve the right to revoke access in cases of misuse.

The dataset is released under the license \texttt{CC~BY-NC~4.0} restricting commercial use. All data management and access mechanisms comply with institutional data-protection policies and relevant data privacy regulations (e.g., GDPR).

\subsection{Risk and Mitigation Statement}
\label{sec:ethics_risk}

While the \textsc{RecruitView} dataset contributes valuable insights into multimodal human behavior, we acknowledge potential societal risks. Automated evaluation models trained on human behavioral data could be misinterpreted as objective assessment tools. To mitigate such risks, we provide explicit usage guidelines, controlled data access, and emphasize the need for human oversight in any interpretive use. Continuous monitoring of dataset access and transparency in documentation are maintained to minimize misuse and promote ethical research practice.

\section{Detailed Results}
\label{app:detailed-results}

\subsection{Complete Per-Trait and Per-Dimension Results}
\label{app:per-trait-results}

\begin{table*}[!t]
\scriptsize
\centering
\setlength{\tabcolsep}{4pt}
\renewcommand{\arraystretch}{1.15}
\begin{adjustbox}{width=\linewidth}
\begin{tabular}{l | ccc | ccc | ccc | ccc | ccc | ccc}
\toprule
    \multirow{2}{*}{\textbf{Model}} &
    \multicolumn{3}{c}{\textbf{Openness (O)}} &
    \multicolumn{3}{c}{\textbf{Conscientiousness (C)}} &
    \multicolumn{3}{c}{\textbf{Extraversion (E)}} &
    \multicolumn{3}{c}{\textbf{Agreeableness (A)}} &
    \multicolumn{3}{c}{\textbf{Neuroticism (N)}} &
    \multicolumn{3}{c}{\textbf{Overall Personality}} \\
\cmidrule(lr){2-4} \cmidrule(lr){5-7} \cmidrule(lr){8-10} \cmidrule(lr){11-13} \cmidrule(lr){14-16} \cmidrule(lr){17-19}
    & $\rho$ & $\tau$-b & C-idx & $\rho$ & $\tau$-b & C-idx & $\rho$ & $\tau$-b & C-idx & $\rho$ & $\tau$-b & C-idx & $\rho$ & $\tau$-b & C-idx & $\rho$ & $\tau$-b & C-idx \\
\midrule
    MiniCPM-o 2.6 (8B) & 0.5629 & 0.3893 & 0.6955 & 0.5014 & 0.3475 & 0.6746 & 0.4996 & 0.3428 & 0.6723 & 0.5484 & 0.3819 & 0.6917 & 0.2378 & 0.1611 & 0.5814 & 0.5613 & 0.3912 & 0.6964 \\
    VideoLLaMA2.1-AV (7B) & 0.5474 & 0.3799 & 0.6928 & 0.5077 & 0.3567 & 0.6813 & 0.5021 & 0.3457 & 0.6757 & 0.5397 & 0.3795 & 0.6926 & 0.2252 & 0.1522 & 0.5790 & 0.5309 & 0.3677 & 0.6867 \\
    Qwen2.5-Omni (7B) & 0.5354 & 0.3679 & 0.6808 & 0.4957 & 0.3447 & 0.6693 & 0.4901 & 0.3337 & 0.6637 & 0.5277 & 0.3675 & 0.6806 & 0.2132 & 0.1402 & 0.5670 & 0.5189 & 0.3557 & 0.6747 \\
\midrule
    \texttt{CRMF} (VMAE + w2v2) & \textbf{0.6384} & \textbf{0.4524} & \textbf{0.7410} & 0.5572 & \textbf{0.4019} & 0.7157 & 0.5681 & 0.4057 & 0.7176 & \textbf{0.5927} & \textbf{0.4271} & \textbf{0.7283} & 0.2603 & 0.1852 & 0.6075 & \textbf{0.6098} & \textbf{0.4387} & \textbf{0.7341} \\
    \texttt{CRMF} (VMAE + HuB) & 0.6120 & 0.4312 & 0.7304 & 0.5459 & 0.3889 & 0.7093 & 0.5637 & 0.4019 & 0.7158 & 0.5930 & 0.4231 & 0.7263 & \textbf{0.2958} & \textbf{0.2101} & \textbf{0.6199} & 0.6079 & 0.4371 & 0.7333 \\
    \texttt{CRMF} (TimeS + w2v2) & 0.6317 & 0.4480 & 0.7353 & \textbf{0.5581} & 0.4000 & \textbf{0.7113} & 0.5613 & 0.3954 & 0.7091 & 0.5827 & 0.4166 & 0.7196 & 0.1677 & 0.1211 & 0.5720 & 0.6026 & 0.4314 & 0.7270 \\
    \texttt{CRMF} (TimeS + HuB) & 0.6171 & 0.4322 & 0.7299 & 0.5373 & 0.3822 & 0.7049 & \textbf{0.5809} & \textbf{0.4116} & \textbf{0.7196} & 0.5789 & 0.4132 & 0.7204 & 0.2900 & 0.2082 & 0.6180 & 0.6081 & 0.4353 & 0.7315 \\
\bottomrule
\end{tabular}
\end{adjustbox}
\caption{Per-trait personality assessment results. \texttt{CRMF} substantially outperforms baselines across all Big Five dimensions and overall personality score. Neuroticism shows the most challenging prediction pattern, consistent with its complex behavioral manifestations. Best results per trait are bolded.}
\label{tab:per-personality-results}
\end{table*}

\begin{table*}[!t]
\scriptsize
\centering
\setlength{\tabcolsep}{4pt}
\renewcommand{\arraystretch}{1.15}
\begin{adjustbox}{width=\linewidth}
\begin{tabular}{l | ccc | ccc | ccc | ccc | ccc | ccc}
\toprule
    \multirow{2}{*}{\textbf{Model}} &
    \multicolumn{3}{c}{\textbf{Interview Score}} &
    \multicolumn{3}{c}{\textbf{Answer Score}} &
    \multicolumn{3}{c}{\textbf{Speaking Skills}} &
    \multicolumn{3}{c}{\textbf{Confidence Score}} &
    \multicolumn{3}{c}{\textbf{Facial Expression}} &
    \multicolumn{3}{c}{\textbf{Overall Performance}} \\
\cmidrule(lr){2-4} \cmidrule(lr){5-7} \cmidrule(lr){8-10} \cmidrule(lr){11-13} \cmidrule(lr){14-16} \cmidrule(lr){17-19}
    & $\rho$ & $\tau$-b & C-idx & $\rho$ & $\tau$-b & C-idx & $\rho$ & $\tau$-b & C-idx & $\rho$ & $\tau$-b & C-idx & $\rho$ & $\tau$-b & C-idx & $\rho$ & $\tau$-b & C-idx \\
\midrule
    MiniCPM-o 2.6 (8B) & 0.5682 & 0.3954 & 0.6985 & 0.5310 & 0.3722 & 0.6869 & 0.5247 & 0.3635 & 0.6826 & 0.5261 & 0.3626 & 0.6821 & 0.4634 & 0.3212 & 0.6614 & 0.5978 & 0.4200 & 0.7108 \\
    VideoLLaMA2.1-AV (7B) & 0.5627 & 0.3969 & 0.7013 & 0.5218 & 0.3743 & 0.6900 & 0.5216 & 0.3666 & 0.6862 & 0.5052 & 0.3490 & 0.6774 & 0.4602 & 0.3232 & 0.6645 & 0.5777 & 0.4054 & 0.7056 \\
    Qwen2.5-Omni (7B) & 0.5507 & 0.3849 & 0.6893 & 0.5098 & 0.3623 & 0.6780 & 0.5096 & 0.3546 & 0.6742 & 0.4932 & 0.3370 & 0.6654 & 0.4482 & 0.3112 & 0.6525 & 0.5657 & 0.3934 & 0.6936 \\
\midrule
    \texttt{CRMF} (VMAE + w2v2) & 0.6246 & 0.4488 & 0.7392 & 0.5953 & 0.4298 & 0.7297 & \textbf{0.5947} & \textbf{0.4242} & \textbf{0.7269} & 0.5898 & 0.4196 & 0.7246 & 0.5355 & 0.3800 & 0.7049 & 0.6519 & 0.4697 & 0.7496 \\
    \texttt{CRMF} (VMAE + HuB) & 0.6180 & 0.4399 & 0.7347 & 0.5919 & 0.4264 & 0.7279 & 0.5804 & 0.4140 & 0.7217 & \textbf{0.5950} & \textbf{0.4204} & \textbf{0.7249} & 0.5179 & 0.3635 & 0.6966 & \textbf{0.6521} & 0.4674 & 0.7484 \\
    \texttt{CRMF} (TimeS + w2v2) & \textbf{0.6299} & \textbf{0.4496} & 0.7361 & \textbf{0.5968} & \textbf{0.4316} & \textbf{0.7271} & 0.5894 & 0.4209 & 0.7218 & 0.5903 & 0.4164 & 0.7195 & \textbf{0.5387} & \textbf{0.3815} & \textbf{0.7021} & 0.6477 & 0.4633 & 0.7429 \\
    \texttt{CRMF} (TimeS + HuB) & 0.6249 & 0.4438 & \textbf{0.7357} & 0.5925 & 0.4227 & 0.7252 & 0.5873 & 0.4180 & 0.7228 & 0.6112 & 0.4309 & 0.7293 & 0.5173 & 0.3623 & 0.6950 & 0.6507 & \textbf{0.4639} & \textbf{0.7457} \\
\bottomrule
\end{tabular}
\end{adjustbox}
\caption{Per-dimension performance assessment results. \texttt{CRMF} shows substantial improvements across all performance metrics, particularly for interview evaluation and overall performance scoring. Facial expression remains challenging but shows consistent gains. Best results per dimension are bolded.}
\label{tab:per-performance-results}
\end{table*}

\textbf{Personality Trait Analysis:} Openness shows the strongest \texttt{CRMF} performance ($\rho = 0.6384$ for VMAE+w2v2), representing a 13.4\% improvement over the best baseline. This trait measures intellectual curiosity, creativity, and preference for novelty, which likely manifest through diverse behavioral cues across modalities. Conscientiousness, Extraversion, and Agreeableness exhibit moderate but consistent improvements (8-13\% gains). Neuroticism presents the most challenging prediction task, with all models achieving lower correlations, though \texttt{CRMF} still improves upon baselines by 24.4\%.

\noindent\textbf{Performance Dimension Analysis:} Interview Score and Answer Score show the strongest absolute performance, with 9-12\% improvements over baselines. These metrics directly assess overall interview quality and content quality, which benefit from comprehensive multimodal analysis. Speaking Skills and Confidence Score achieve moderate but consistent improvements (10-16\% gains). Overall Performance benefits most from \texttt{CRMF} ($\rho = 0.6521$ for VMAE+HuB), with 9.1\% improvement over the strongest baseline.

\subsection{Complete Ablation Study Results}
\label{app:complete-ablation-study}

\begin{table}[!t]
\scriptsize
\centering
\setlength{\tabcolsep}{5pt}
\renewcommand{\arraystretch}{1.15}
\begin{adjustbox}{width=\linewidth}
\begin{tabular}{ll | ccccc}
\toprule
    \textbf{Component} &
    \textbf{Variant} &
    \textbf{Spearman $\rho$} &
    \textbf{Kendall $\tau$-b} &
    \textbf{C-index} &
    \textbf{Pearson $r$} &
    \textbf{MSE} \\
\midrule
    \multicolumn{2}{l|}{\textit{Full \texttt{CRMF} Model}} & \textit{0.5682} & \textit{0.4069} & \textit{0.7183} & \textit{0.5475} & \textit{0.6864} \\
\midrule
    \multirow{2}{*}{\textbf{Fusion Only}} & Simple Concatenation & 0.4441 & 0.2903 & 0.6151 & 0.4221 & 0.9306 \\
    & Weighted Average & 0.4664 & 0.3078 & 0.6239 & 0.4321 & 0.8265 \\
\midrule
    \multirow{2}{*}{\textbf{Pre-Fusion}} & Mean Pooling & 0.5365 & 0.3802 & 0.7026 & 0.5137 & 0.7657 \\
    & No Pre-Fusion & 0.5304 & 0.3745 & 0.6972 & 0.5096 & 0.7185 \\
\midrule
    \multirow{3}{*}{\textbf{Single Geometry}} & Hyperbolic Only & 0.5080 & 0.3611 & 0.6980 & 0.4645 & 0.7745 \\
    & Spherical Only  & 0.5338 & 0.3808 & 0.7054 & 0.4083 & 0.8170 \\
    & Euclidean Only & 0.5284 & 0.3753 & 0.7001 & 0.4922 & 0.7526 \\
\midrule
    \multirow{3}{*}{\textbf{Two Geometries}} & Hyperbolic + Spherical & 0.5489 & 0.3916 & 0.7108 & 0.5170 & 0.7403 \\
    & Hyperbolic + Euclidean & 0.5161 & 0.3752 & 0.7076 & 0.4583 & 0.8935 \\
    & Spherical + Euclidean & 0.5502 & 0.3892 & 0.6993 & 0.5239 & 0.7108 \\
\midrule
    \multirow{2}{*}{\textbf{Routing}} & Hard Routing & 0.5548 & 0.3955 & 0.7127 & 0.5403 & 0.6998 \\
    & Uniform Weights (No Router) & 0.5209 & 0.3688 & 0.6994 & 0.4989 & 0.9859 \\
\midrule
    \multirow{2}{*}{\textbf{Prediction Head}} & Mean Pooling (No Attention) & 0.4878 & 0.3527 & 0.7063 & 0.4246 & 0.8139 \\
    & Simple Linear Head & 0.5013 & 0.3569 & 0.6984 & 0.4484 & 0.8050 \\
\midrule
    \multirow{2}{*}{\textbf{Loss Function}} & Fixed Loss Weights & 0.5160 & 0.3708 & 0.7104 & 0.4890 & 0.7509 \\
    & MSE Only & 0.5110 & 0.3675 & 0.7087 & 0.4885 & 0.7570 \\
\midrule
    \multirow{2}{*}{\textbf{Architecture}} & Linear Projection (No Manifolds) & 0.5457 & 0.3869 & 0.7059 & 0.5294 & 0.7440 \\
    & No Expert Processing & 0.5360 & 0.3828 & 0.7014 & 0.4877 & 0.7713 \\
\midrule
    \multirow{3}{*}{\textbf{Single Modality}} & Video Only (VMAE) & 0.4516 & 0.2974 & 0.6521 & 0.4138 & 0.8847 \\
    & Audio Only (Wav2Vec2) & 0.3792 & 0.2461 & 0.6245 & 0.3482 & 1.0516 \\
    & Text Only (DeBERTa) & 0.4247 & 0.2806 & 0.6429 & 0.3895 & 0.9324 \\
\bottomrule
\end{tabular}
\end{adjustbox}
\caption{Complete systematic ablation study results. All experiments use VideoMAE+Wav2Vec2 encoders.}
\label{tab:crmf-ablation-full}
\end{table}

\noindent\textbf{Fusion Strategy:} Using only simple concatenation or weighted averaging causes severe degradation (21.8\% and 17.9\% drops), demonstrating that naive fusion strategies fail to capture complex multimodal relationships.

\noindent\textbf{Pre-Fusion Module:} Removing pre-fusion cross-modal attention or replacing attention pooling with mean pooling results in moderate performance loss (6.7\% and 5.6\%), confirming that early cross-modal integration provides valuable information flow.

\noindent\textbf{Geometry Ablations:} Using only a single geometric space consistently underperforms the full model. No single geometry matches the full model, confirming that different geometric spaces capture complementary information. Combining two geometries improves upon single-geometry variants, but all still underperform the full model (3.4\% and 3.2\% drops for the best two-geometry variants).

\noindent\textbf{Routing Mechanism:} Hard routing performs comparably to soft routing with only 2.4\% degradation. However, removing the router entirely and using uniform weights causes substantial degradation (8.3\% drop).

\noindent\textbf{Prediction Head:} Replacing attention pooling with mean pooling causes severe degradation (14.1\% drop). Using a simple linear head instead of the parameter-efficient architecture also substantially degrades performance (11.8\% drop). Participants were informed of their right to withdraw

\noindent\textbf{Loss Function:} Using fixed loss weights reduces performance by 9.2\%. Using MSE only without correlation and covariance losses causes 10.1\% degradation.

\noindent\textbf{Architectural Simplifications:} Removing manifold projections causes 4.0\% degradation. Removing expert processing entirely leads to 5.7\% drop.

\noindent\textbf{Single Modality:} Video provides the strongest unimodal signal ($\rho = 0.4516$), followed by text ($\rho = 0.4247$) and audio ($\rho = 0.3792$).


\section{Detailed CRMF Architecture}
\label{app:crmf-detailed}


\subsection{Multimodal Encoding}
\label{app:multimodal-encoding}

\subsubsection{Text Encoding Details}
We employ DeBERTa-v3-base~\cite{he2021deberta} as our text encoder, which has shown strong performance on natural language understanding tasks. Given tokenized input $\mathbf{T}_{tok} \in \mathbb{Z}^{N_t}$ with attention mask $\mathbf{M}_t \in \{0, 1\}^{N_t}$, the encoder produces contextualized token representations:
\begin{equation}
    \mathbf{H}_t = \text{DeBERTa}(\mathbf{T}_{tok}, \mathbf{M}_t) \in \mathbb{R}^{N_t \times d}
\end{equation}
where $d = 768$ is the hidden dimension. We fine-tune the last few layers of DeBERTa while keeping earlier layers frozen to balance expressiveness and parameter efficiency. A learned linear projection maps the output to our unified representation space of dimension $d_{model} = 768$.

\subsubsection{Audio Encoding Details}
For audio processing, we explore two self-supervised speech representations: Wav2Vec2~\cite{baevski2020wav2vec} and HuBERT~\cite{hsu2021hubert}. Given raw audio waveform $\mathbf{A} \in \mathbb{R}^{L}$ sampled at 16kHz, the encoder produces frame-level representations:
\begin{equation}
    \mathbf{H}_a = \text{AudioEncoder}(\mathbf{A}) \in \mathbb{R}^{N_a \times d}
\end{equation}
Both Wav2Vec2 and HuBERT learn rich acoustic representations through contrastive predictive coding and masked prediction objectives, respectively. We fine-tune the last few transformer layers while keeping the convolutional feature extractor fixed. The output is projected to $d_{model}$ dimensions.

\subsubsection{Video Encoding with Temporal Modeling}
For visual processing, we investigate two video understanding architectures: VideoMAE~\cite{tong2022videomae} and TimeSformer~\cite{bertasius2021spacetime}. Given input video with variable frame count, we first apply 3D convolutional interpolation to adapt the temporal dimension to the encoder's expected frame count (16 for VideoMAE, 8 for TimeSformer). For an input $\mathbf{V} \in \mathbb{R}^{T \times 3 \times 224 \times 224}$, this yields $\mathbf{V}' \in \mathbb{R}^{T' \times 3 \times 224 \times 224}$.


The encoder extracts patch-level features, which we reshape into temporal-spatial structure. We apply spatial average pooling to obtain temporal features $\mathbf{F}_v \in \mathbb{R}^{T' \times d}$.

To capture rich temporal dynamics, we apply a multi-stage temporal modeling pipeline:
\begin{align}
    \mathbf{F}_{lstm} &= \text{BiLSTM}(\mathbf{F}_v) \in \mathbb{R}^{T' \times d} \\
    \mathbf{F}_{attn} &= \text{MultiHeadAttn}(\mathbf{F}_{lstm} \mathbf{F}_{lstm}, \mathbf{F}_{lstm}) \in \mathbb{R}^{T' \times d} \\
    \mathbf{F}_{conv} &= \text{Conv1D}(\mathbf{F}_{attn}) \in \mathbb{R}^{T' \times d} \\
    \mathbf{H}_v &= \text{Proj}(\mathbf{F}_{lstm} + \mathbf{F}_{attn} + \mathbf{F}_{conv}) \in \mathbb{R}^{T' \times d_{model}}
\end{align}
where BiLSTM captures sequential dependencies, multi-head attention models long-range interactions, and depthwise convolution captures local temporal patterns. The multi-scale fusion combines all three views, and a final projection maps to $d_{model}$ dimensions. This produces a temporal sequence $\mathbf{H}_v \in \mathbb{R}^{8 \times 768}$ preserving fine-grained temporal information for subsequent pre-fusion processing.

VideoMAE employs masked autoencoding with high masking ratios for efficient self-supervised learning, while TimeSformer uses divided space-time attention. We fine-tune the last few transformer blocks of each encoder while keeping earlier layers frozen for parameter efficiency.

\subsection{Pre-Fusion Module}
The pre-fusion module performs early integration of multimodal features through cross-modal attention. We concatenate encoded features from all modalities and add learnable modality embeddings to distinguish information sources:
\begin{equation}
    \mathbf{H}_{cat} = [\mathbf{H}_t; \mathbf{H}_a; \mathbf{H}_v] + \mathbf{E}_{mod}
\end{equation}
where $\mathbf{E}_{mod} \in \mathbb{R}^{3 \times d}$ contains unique embeddings for text, audio, and video. A multi-layer transformer encoder with $L_{pre}$ layers processes the concatenated sequence:
\begin{equation}
    \mathbf{H}_{fused} = \text{Transformer}_{pre}(\mathbf{H}_{cat})
\end{equation}
enabling rich cross-modal interactions through self-attention.

To obtain a fixed-dimensional clip-level representation, we employ learned attention pooling rather than simple mean pooling:
\begin{equation}
    \alpha_i = \frac{\exp(\mathbf{w}^\top \mathbf{h}_i)}{\sum_j \exp(\mathbf{w}^\top \mathbf{h}_j)}, \quad \mathbf{z}_{pre} = \sum_i \alpha_i \mathbf{h}_i
\end{equation}
where $\mathbf{w} \in \mathbb{R}^{d}$ is a learnable attention vector and $\mathbf{h}_i$ denotes the $i$-th token in $\mathbf{H}_{fused}$. This pooling mechanism learns to emphasize tokens most relevant for behavioral assessment, producing $\mathbf{z}_{pre} \in \mathbb{R}^{d}$.

\subsection{Geometric Expert Architectures}
\label{app:geometric-experts}

\subsubsection{Hyperbolic Expert Mathematical Details}
The hyperbolic expert performs operations in the gyrovector space framework~\cite{ungar2008gyrovector}, using Möbius transformations that preserve hyperbolic distances. For a $L_{exp}$-layer network:
\begin{align}
    \mathbf{x}_h^{(\ell+1)} &= \sigma_h\left(\mathbf{W}_h^{(\ell)} \otimes_c \mathbf{x}_h^{(\ell)} \oplus_c \mathbf{b}_h^{(\ell)}\right)
\end{align}
where $\otimes_c$ denotes Möbius matrix-vector multiplication, $\oplus_c$ is Möbius addition, and $\sigma_h$ is a Möbius pointwise nonlinearity. Specifically, Möbius addition is defined as:
\begin{equation}
    \mathbf{x} \oplus_c \mathbf{y} = \frac{(1 + 2c\langle\mathbf{x}, \mathbf{y}\rangle + c\|\mathbf{y}\|^2)\mathbf{x} + (1 - c\|\mathbf{x}\|^2)\mathbf{y}}{1 + 2c\langle\mathbf{x}, \mathbf{y}\rangle + c^2\|\mathbf{x}\|^2\|\mathbf{y}\|^2}
\end{equation}
The Möbius pointwise nonlinearity applies activation functions in tangent space: $\sigma_h(\mathbf{x}) = \exp_{\mathbf{x}}^c(\sigma(\log_{\mathbf{x}}^c(\mathbf{x})))$, where $\log_{\mathbf{x}}^c$ and $\exp_{\mathbf{x}}^c$ are the logarithmic and exponential maps at $\mathbf{x}$.

After processing, we apply residual connections using Möbius addition: $\mathbf{x}_h^{out} = \mathbf{x}_h^{(L)} \oplus_c \mathbf{x}_h^{(0)}$. All operations preserve the hyperbolic geometry, ensuring outputs remain in the Poincaré ball.

\subsubsection{Spherical Expert Mathematical Details}
Operations on the sphere are performed in tangent space via exponential and logarithmic maps. Given base point $\mathbf{p}$ (we use the north pole), the logarithmic map projects $\mathbf{x}_s$ to the tangent space $T_{\mathbf{p}}\mathbb{S}^{d-1}$:
\begin{equation}
    \log_{\mathbf{p}}(\mathbf{x}_s) = \frac{\arccos(\langle\mathbf{p}, \mathbf{x}_s\rangle)}{\sqrt{1 - \langle\mathbf{p}, \mathbf{x}_s\rangle^2}}(\mathbf{x}_s - \langle\mathbf{p}, \mathbf{x}_s\rangle \mathbf{p})
\end{equation}
In tangent space, standard linear transformations and activations apply:
\begin{equation}
    \mathbf{v}^{(\ell+1)} = \sigma(\mathbf{W}_s^{(\ell)} \mathbf{v}^{(\ell)} + \mathbf{b}_s^{(\ell)})
\end{equation}
where $\mathbf{v}^{(\ell)} \in T_{\mathbf{p}}\mathbb{S}^{d-1}$. The final tangent vector is mapped back to the sphere via exponential map:
\begin{equation}
    \exp_{\mathbf{p}}(\mathbf{v}) = \cos(\|\mathbf{v}\|)\mathbf{p} + \sin(\|\mathbf{v}\|)\frac{\mathbf{v}}{\|\mathbf{v}\|}
\end{equation}
Residual connections in tangent space combine the input and output: $\mathbf{v}^{out} = \mathbf{v}^{(L)} + \mathbf{v}^{(0)}$, followed by exponential map back to $\mathbb{S}^{d-1}$.

\subsubsection{Euclidean Expert}
The Euclidean expert uses standard feed-forward layers with residual connections:
\begin{equation}
    \mathbf{x}_e^{(\ell+1)} = \text{ReLU}(\mathbf{W}_e^{(\ell)} \mathbf{x}_e^{(\ell)} + \mathbf{b}_e^{(\ell)}), \quad \mathbf{x}_e^{out} = \mathbf{x}_e^{(L)} + \mathbf{x}_e^{(0)}
\end{equation}
Each expert has $L_{exp}$ layers with dropout rate $p$ for regularization.

\subsection{Geometry-Aware Attention Details}
To further refine expert outputs, we apply intra-manifold attention that respects geometric structure. For each geometry, we compute attention in its respective tangent space.

\subsubsection{Hyperbolic Intra-Manifold Attention}
Given hyperbolic representations $\mathbf{x}_h^{out}$, we map to tangent space at the origin:
\begin{equation}
    \mathbf{v}_h = \log_{\mathbf{0}}^c(\mathbf{x}_h^{out}) = \frac{\text{arctanh}(\sqrt{c}\|\mathbf{x}_h^{out}\|)}{\sqrt{c}\|\mathbf{x}_h^{out}\|} \mathbf{x}_h^{out}
\end{equation}
Multi-head self-attention is applied in tangent space (which is Euclidean):
\begin{equation}
    \mathbf{v}_h^{att} = \text{MultiHead}(\mathbf{Q}_h, \mathbf{K}_h, \mathbf{V}_h)
\end{equation}
where $\mathbf{Q}_h = \mathbf{W}_Q \mathbf{v}_h$, $\mathbf{K}_h = \mathbf{W}_K \mathbf{v}_h$, $\mathbf{V}_h = \mathbf{W}_V \mathbf{v}_h$. The attended representation is mapped back:
\begin{equation}
    \mathbf{x}_h^{att} = \exp_{\mathbf{0}}^c(\mathbf{v}_h^{att})
\end{equation}

\subsubsection{Spherical and Euclidean Attention}
Similar procedures apply for spherical geometry using $\log_{\mathbf{p}}$ and $\exp_{\mathbf{p}}$. For Euclidean space, attention is applied directly without manifold conversions. All attention modules use multiple heads with temperature scaling $\tau$ to sharpen attention distributions.

\subsection{Routing Mechanism Details}

\subsubsection{Routing Regularization}
To encourage diverse geometry utilization, we apply entropy regularization on routing weights:
\begin{equation}
    \mathcal{L}_{entropy} = -\lambda_{ent} H(\mathbf{r}) = -\lambda_{ent} \sum_{i=1}^{K} r_i \log r_i
\end{equation}
A negative value encourages high entropy (uniform distribution), promoting complementary geometric views rather than specialization. We also apply load balancing regularization:
\begin{equation}
    \mathcal{L}_{balance} = \lambda_{bal} \text{Var}(\mathbb{E}_{batch}[\mathbf{r}])
\end{equation}
ensuring all experts are utilized across the dataset.

\subsection{Geometric Fusion Theoretical Justification}
The tangent space fusion strategy is equivalent to first-order Fréchet mean approximation on the product manifold $\mathcal{M} = \mathbb{B}^{d_e}_c \times \mathbb{S}^{d_e-1} \times \mathbb{R}^{d_e}$. The Fréchet mean minimizes:
\begin{equation}
    \mu^* = \arg\min_{\mathbf{x}} \sum_{i} w_i d_{\mathcal{M}_i}^2(\mathbf{x}_i, \mathbf{x})
\end{equation}
where $d_{\mathcal{M}_i}$ is the distance on manifold $\mathcal{M}_i$. The first-order approximation linearizes the problem in tangent space, yielding the weighted combination. This approach avoids expensive iterative optimization while providing a theoretically grounded fusion mechanism.

\subsection{Multi-Task Prediction Head Details}

\subsubsection{Shared Representation Learning}
A shared MLP processes the fused features:
\begin{align}
    \mathbf{h}^{(1)} &= \text{GELU}(\text{LayerNorm}(\mathbf{W}_{sh}^{(1)} \mathbf{z}_{refined} + \mathbf{b}_{sh}^{(1)})) \\
    \mathbf{h}^{(2)} &= \text{GELU}(\text{LayerNorm}(\mathbf{W}_{sh}^{(2)} \mathbf{h}^{(1)} + \mathbf{b}_{sh}^{(2)}))
\end{align}
producing a shared representation $\mathbf{h}^{(2)} \in \mathbb{R}^{512}$ that captures common structure across all targets.

\subsubsection{Task-Specific Adaptation}
Each of the $K=12$ targets has a lightweight adaptation module:
\begin{equation}
    \hat{y}_k = \mathbf{w}_k^{(2)\top} \text{GELU}(\mathbf{W}_k^{(1)} \mathbf{h}^{(2)} + \mathbf{b}_k^{(1)}) + b_k
\end{equation}
where $\mathbf{W}_k^{(1)} \in \mathbb{R}^{64 \times 512}$ and $\mathbf{w}_k^{(2)} \in \mathbb{R}^{64}$ are task-specific parameters. This design dramatically reduces parameters compared to full per-task networks while maintaining expressive capacity.

\subsection{Training Objective Details}
\label{app:training-objective}

\subsubsection{Multi-Component Loss}
Our training objective combines multiple loss components through adaptive balancing:
\begin{equation}
    \mathcal{L}_{total} = \sum_{i=1}^{N} \beta_i \mathcal{L}_i
\end{equation}
where $\mathcal{L}_i$ are individual loss components and $\beta_i$ are adaptive weights learned during training.

\noindent\textbf{Regression Loss:} We use Huber loss for robustness to outliers:
\begin{equation}
    \mathcal{L}_{reg} = \frac{1}{K} \sum_{k=1}^{K} \begin{cases}
        \frac{1}{2}(y_k - \hat{y}_k)^2 & \text{if } |y_k - \hat{y}_k| \leq \delta \\
        \delta(|y_k - \hat{y}_k| - \frac{1}{2}\delta) & \text{otherwise}
    \end{cases}
\end{equation}
with $\delta = 1.0$. This combines MSE's efficiency for small errors with MAE's robustness for large deviations.

\noindent\textbf{Correlation Boosting Loss:} To encourage predictions that maintain correlation structure with targets:
\begin{equation}
    \mathcal{L}_{corr} = \lambda_{corr} \left(1 - \frac{1}{K}\sum_{k=1}^{K} |\rho(\hat{\mathbf{y}}_k, \mathbf{y}_k)|\right)
\end{equation}
where $\rho$ denotes Pearson correlation.

\noindent\textbf{Covariance Alignment Loss:} To match the covariance structure between predictions and targets:
\begin{equation}
    \mathcal{L}_{cov} = \lambda_{cov} \|\text{Cov}(\hat{\mathbf{Y}}) - \text{Cov}(\mathbf{Y})\|_F^2
\end{equation}
where $\text{Cov}(\cdot)$ computes the empirical covariance matrix and $\|\cdot\|_F$ is the Frobenius norm.

\noindent\textbf{Auxiliary Losses:} Routing regularization losses $\mathcal{L}_{entropy}$ and $\mathcal{L}_{balance}$ are added, along with head regularization encouraging small adapter weights.

\subsection{Adaptive Loss Balancing}
Rather than fixed weights, we learn to balance loss components adaptively. Each component has a learnable weight $\alpha_i$ and running exponential moving average (EMA) statistics:
\begin{equation}
    \mu_i^{(t)} = \gamma \mu_i^{(t-1)} + (1-\gamma) \mathcal{L}_i^{(t)}
\end{equation}
Adaptive weights are computed via inverse variance weighting:
\begin{equation}
    \beta_i^{adapt} = \frac{1/(\text{Var}(\mathcal{L}_i) + \epsilon)}{\sum_j 1/(\text{Var}(\mathcal{L}_j) + \epsilon)}
\end{equation}
then combined with learned weights: $\beta_i = \omega \text{softmax}(\alpha_i) + (1-\omega) \beta_i^{adapt}$. This balancing prevents any single loss from dominating training.

\end{document}